\pdfoutput=1

\documentclass[11pt]{article}

\usepackage[preprint]{acl}

\usepackage{times}
\usepackage{latexsym}

\usepackage[T1]{fontenc}

\usepackage[utf8]{inputenc}

\usepackage{microtype}

\usepackage{inconsolata}

\usepackage{graphicx}

\usepackage{ragged2e,float}
\usepackage{graphicx,hyperref}
\usepackage{amsmath,amsthm,amssymb}
\usepackage{bbm}
\usepackage{booktabs,multicol,multirow}
\usepackage[]{todonotes} 
\usepackage{algorithm,algorithmic}
\usepackage{comment}
\theoremstyle{definition}
\theoremstyle{remark}

\usepackage{soul}

\usepackage{mathrsfs}
\usepackage{pifont}
\newcommand{\cmark}{\ding{51}}
\newcommand{\xmark}{\ding{55}}
\newcommand{\flower}{\ding{95}}
\usepackage{cleveref}

\title{Factual Inconsistency in Data-to-Text Generation Scales Exponentially with LLM Size: A Statistical Validation}

\author{
        Joy Mahapatra \and Soumyajit Roy \and Utpal Garain \\
        Indian Statistical Institute Kolkata \\
        \small{\textbf{Correspondence:} \href{mailto:joymahapatra90@gmail.com}{joymahapatra90@gmail.com}}
        }

\begin{document}
\maketitle

\begin{abstract}
Monitoring factual inconsistency is essential for ensuring trustworthiness in data-to-text generation (D2T).
While large language models (LLMs) have demonstrated exceptional performance across various D2T tasks, previous studies on scaling laws have primarily focused on generalization error through power law scaling to LLM size (i.e., the number of model parameters).
However, no research has examined the impact of LLM size on factual inconsistency in D2T.
In this paper, we investigate how factual inconsistency in D2T scales with LLM size by exploring two scaling laws: power law and exponential scaling.
To rigorously evaluate and compare these scaling laws, we employ a statistical validation framework consisting of three key stages: predictive performance estimation, goodness-of-fit assessment, and comparative analysis.
For a comprehensive empirical study, we analyze three popular LLM families across five D2T datasets, measuring factual inconsistency inversely using four state-of-the-art consistency metrics.
Our findings, based on exhaustive empirical results and validated through our framework, reveal that, contrary to the widely assumed power law scaling, factual inconsistency in D2T follows an exponential scaling with LLM size.
\end{abstract}

\section{Introduction}
\label{sec:intro}
\begin{figure}[H]
    \centering
    \includegraphics[width=\linewidth]{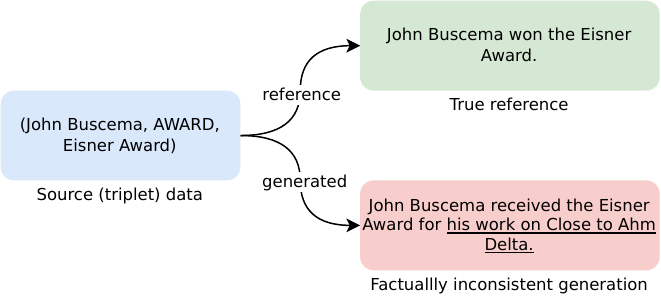}
    \caption{Example of data-to-text generation from the DART dataset, with a factually inconsistent output from the Pythia-1.4B model.}
    \label{fig:example}
\end{figure}

Data-to-text (D2T) generation~\citep{lin2024survey,li2024unifying} converts semi-structured data (e.g., tables) into natural language, with applications in conversation systems, automated journalism, and other fields.
A key challenge in D2T is factual inconsistency~\citep{li2022faithfulness,huang2023survey}---when generated text fails to entails with input facts---leading to hallucinations that undermine trust in D2T models (\autoref{fig:example}).
Therefore, it is essential to monitor and mitigate factual inconsistency in order to construct trustworthy D2T models.

Large language models (LLMs) have achieved remarkable success in D2T, primarily due to their massive model sizes (parameter counts) and training on vast text corpora~\citep{lorandi2024high}.
Several studies shows that LLMs often adhere to scaling laws, typically power laws, governing generalization error or perplexity in relation to model size~\citep{kaplan2020scaling,hoffmann2022training}.
These scaling laws play a crucial role in predicting model performance, guiding hyperparameter tuning, estimating computational costs, and optimizing resource allocation~\citep{hendrycksforthcomingintroduction,zhang2024when}.
Existing LLM scaling laws in D2T focus on generalization loss or test perplexity~\citep{bahri2024explaining}, overlooking factual inconsistency.
Understanding how factual inconsistency scales with LLM model size can help researchers and practitioners optimize model selection and enhance trustworthiness in D2T, highlighting a key research gap.

In this paper, we address the research gap by examining scaling laws for factual inconsistency in D2T with respect to LLM size.
Unlike prior studies that focus solely on power law scaling, we explore both power law and exponential scaling with a rigorous three-stage statistical validation framework.
This framework comprises three key stages: predictive performance estimation (evaluating Huber loss on held-out data), goodness-of-fit assessment (using an F-test to measure goodness-of-fit), and comparative analysis (utilizing Vuong’s likelihood-ratio test to compare power law and exponential scaling).
By integrating rigorous statistical validation, we ensure more reliable and robust insights, particularly in data-limited settings.
Our study spans three widely used LLM families---Pythia, OPT, and BLOOM---and five well-established D2T datasets: E2E, ViGGO, WebNLG, DART, and WikiTableText.  
Factual inconsistency is quantified as the inverse of factual consistency, measured using four state-of-the-art metrics---\textsc{AlignScore}, \textsc{QAFactEval}, \textsc{SummaC-conv}, and \textsc{UniEval-fact}---which strongly correlate with human judgments.  
Our findings, validated through extensive empirical analysis and the rigorous validation framework, reveal that factual inconsistency in D2T follows exponential scaling with LLM size rather than power law scaling.

\section{Related Work}
\label{sec:background}
\paragraph{Data-to-text generation (D2T) and factual inconsistency.}
Data-to-text generation (D2T)\citep{lin2024survey} aims to transform non-textual, semi-structured data---such as tables, graphs, or slot-value pairs (meaning representation, MR)---into human-readable text.
It can be categorized into three types based on source representation: graph-to-text\citep{gardent2017webnlg,nan2021dart}, table-to-text~\citep{bao2018table}, and meaning representation (MR)-to-text~\citep{novikova2017e2e,juraska2019viggo}.
Recently, LLMs have become foundational models for D2T due to their extensive pre-training on large text datasets~\citep{zhang2022opt} and their high model capacity~\citep{scao2022bloom}.
Moreover, with parameter-efficient fine-tuning techniques~\citep{dettmers2023qlora} and prompt-based learning~\citep{lester2021power}, LLMs have gained widespread popularity for D2T tasks~\citep{raffel2020exploring,lewis2020bart,scao2022bloom}, often outperforming earlier models in generation quality and overall performance~\citep{ge2023openagi}.
In D2T, LLMs are often prone to generating factually inconsistent text, presenting a key research challenge. 
Factual inconsistency, defined as the lack of factual entailment between generated text and input data, contributes to hallucinations and undermines model reliability.
Evaluation methods include human assessment (gold standard but costly) and automatic metrics (scalable but debated).
Recently, trained automatic metrics~\citep{fabbri2022qafacteval,zha2023alignscore} have shown strong correlations with human judgments, making them promising for factual inconsistency evaluation.

\paragraph{Scaling law for LLM.}
Scaling laws for LLMs describe how their performance scales with key factors such as model size (number of parameters) and training data size.  
\citet{hestness2017deep} demonstrated that deep language models follow a power law scaling, laying the foundation for scaling law research.
\citet{kaplan2020scaling} expanded this by systematically analyzing model size, data size, and computational efficiency, reinforcing the dominance of power law scaling in LLM performance. 
As research on scaling laws has expanded, various studies have explored their applications across different task domains, including close-ended text generation~\citep{bansal2022data} and open-ended text generation~\citep{kaplan2020scaling}.  
Recent investigations have further examined scaling in diverse paradigms, such as sparse modeling~\citep{frantar2024scaling} and parameter-efficient fine-tuning~\citep{zhang2024when}.  
Additionally, joint scaling laws---such as additive and multiplicative formulations---are gaining prominence in multi-factor scaling setups~\citep{hoffmann2022training, zhang2024when}.  
Scaling laws offer several key advantages, including optimizing hyperparameter tuning~\citep{hendrycksforthcomingintroduction}, estimating training costs~\citep{haegele2024scaling}, and setting realistic expectations for model performance~\citep{hoffmann2022training}. 
A recent study by \citet{bahri2024explaining} further reinforces the theoretical foundations of scaling laws.

\section{Scaling Law Models and Training}
Moving beyond existing studies, we formulate the scaling law for factual inconsistency in D2T concerning LLM size by considering two models—power law scaling (following a power law function) and exponential scaling (following an exponential function).  
The power law scaling model ($\mathcal{M}_{pow}$) is defined as follows:

\begin{align}
    \mathcal{M}_{pow} : f(x) = 
                \begin{cases}
                    Ax^{\alpha} + B & x \geq 0\\
                    0 & \text{otherwise}
                \end{cases}
\end{align}

Where $A$ and $B$ are case-specific parameters, $\alpha$ is the power law exponent, $x$ represents LLM size, and $f(x)$ denotes factual inconsistency.

Similarly, the exponential scaling model ($\mathcal{M}_{exp}$) is defined as follows:

\begin{align}
    \mathcal{M}_{exp} : F(x) = 
                        \begin{cases}
                            Ce^{\beta x} + D & x \geq 0\\
                            0 & \text{otherwise}
                        \end{cases}
\end{align}

Where $C$ and $D$ are case-specific parameters, $\beta$ is the exponential scaling rate, $x$ represents LLM size, and $f(x)$ denotes factual inconsistency.

The parameters of both models are estimated using maximum likelihood estimation (MLE) on the factual inconsistency score data $\mathcal{D}$, optimized through the standard Huber loss ($\delta = 1$), denoted as $\mathcal{L}_{\text{Huber}}$, due to its robust estimation capability.

\begin{align}
    \hat{A}, \hat{B}, \hat{\alpha} & \leftarrow \operatorname{MLE}(M_{power}, \mathcal{D}, \mathcal{L}_{\text{Huber}})\\
    \hat{C}, \hat{D}, \hat{\beta} & \leftarrow \operatorname{MLE}(M_{exp}, \mathcal{D}, \mathcal{L}_{\text{Huber}})
\end{align}

\section{Statistical Validation Framework}
To empirically study the two scaling models under limited data, we employ a structured three-stage statistical validation framework, consisting of predictive performance estimation, goodness-of-fit assessment, and comparative analysis, as detailed below.

\begin{figure}[H]
    \centering
    \includegraphics[width=1\linewidth]{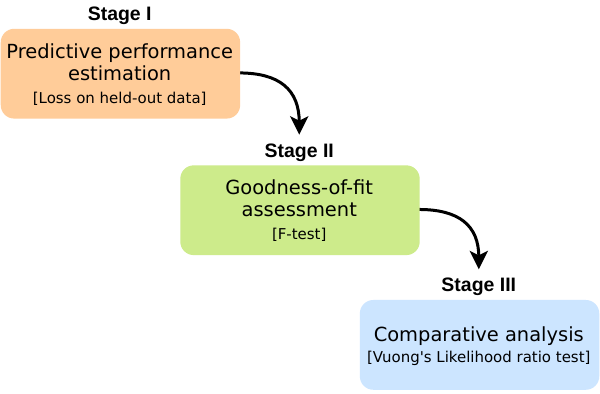}
    \caption{All three stages of our statistical validation framework.}
    \label{fig:validation_stages}
\end{figure}

\begin{itemize}
    \item \textbf{Stage I: Predictive performance estimation.} This validation stage ensures how well the scaling laws generalize in terms of their predictive ability on unseen data.
    To achieve this, we evaluate the scaling laws on held-out data using Huber loss.
    Given the limited data availability, we employ five-fold cross-validation for predictive performance assessment.
    
    \item \textbf{Stage II: Goodness-of-fit assessment.} Predictive performance alone is not sufficient to validate a scaling law; assessing its goodness-of-fit is also crucial for its acceptance.
    Therefore, in this stage, we evaluate the goodness-of-fit of the scaling law models using a goodness-of-fit test---specifically, an F-test for regression~\citep{weisberg2005applied,siegel2016practical}.
    The test statistic for the F-test is calculated as follows:

    \vspace{-3mm}

    \begin{align}
        F_{\text{stat}} &= \frac{SSR_{\mathfrak{R}} - SSR_{\mathfrak{E}}}{df_{\mathfrak{R}} - df_{\mathfrak{E}}} \Big / \frac{SSR_{\mathfrak{E}}}{df_{\mathfrak{E}}}\\
        F_{\text{stat}} &\sim \operatorname{F-distribution}(x)
    \end{align}

    Here, $SSR_{\mathfrak{R}}$ and $SSR_{\mathfrak{E}}$ represent the sum of squared residuals for the reduced and exact models, respectively.
    Similarly, $df_{\mathfrak{R}}$ and $df_{\mathfrak{E}}$ denote the degrees of freedom for the reduced and exact models, respectively.
    We consider our scaling models ($\mathcal{M}_{pow}$ and $\mathcal{M}_{exp}$) as exact models, while the reduced model is represented by a simple mean-response model.
    Since the F-test applies only to linear regression models, we use a log transformation to convert our scaling models into their linear forms.  
    We perform the F-test with a significance level of $p<0.05$, which is often considered a moderate range.  
    If both scaling models qualify in the goodness-of-fit assessment, we proceed to Stage III.
    
    \item \textbf{Stage III: Comparative analysis.} In this final stage of validation, we compare the two scaling law models, $\mathcal{M}{pow}$ and $\mathcal{M}{exp}$, through hypothesis testing to determine which better explains the data.
    Since power law and exponential scaling models are not nested hypotheses, the standard likelihood-ratio test is not applicable.
    Instead, we employ Vuong's likelihood-ratio test~\citep{vuong1989likelihood} for comparison.
    The test statistic for Vuong's likelihood-ratio test is computed as follows:

    \vspace{-3mm}

    \begin{align}
        V_{\text{stat}} &= \frac{\sqrt{n} \cdot \operatorname{mean}(d)}{\sqrt{ \operatorname{Var}(d)}}\\
        V_{\text{stat}} &\sim \operatorname{normal}(0, 1)
    \end{align}

    Where $n$ represents the sample size, and $d$ denotes the $n$-sized sample of the log-likelihood differences between the two scaling law models.
    We conduct Vuong's likelihood ratio test at a stringent significance level of $p<0.005$ to provide highly compelling evidence for our conclusion.
\end{itemize}

\noindent \textbf{Assumptions verification.} In the second and third stages of our validation framework, we incorporate the F-test and Vuong's likelihood-ratio test.  
Both tests rely, directly or indirectly, on the assumption that the residuals of our scaling law models, $\mathcal{M}_{pow}$ and $\mathcal{M}_{exp}$, follow a normal distribution.  
Therefore, validating this assumption is essential.  
To assess the normality of residuals for both $\mathcal{M}_{pow}$ and $\mathcal{M}_{exp}$, we employ the Shapiro–Wilk test in our experiments.

\section{Experiment Setup}
\label{sec:experiment_setup}
\subsection{Dataset}
We utilize five well-known D2T datasets, covering three major D2T types: DART and WebNLG for graph-to-text, WikiTableText for table-to-text, and E2E and ViGGO for MR-to-text.  
All datasets are sourced from~\citep{kasner2023tabgenie} and~\citep{wolf2020transformers}.  
The E2E dataset~\citep{novikova2017e2e,dusek2018findings} contains over 37K MR-text pairs from the restaurant domain, with an average text length of approximately 21 words.  
ViGGO~\citep{juraska2018deep} includes 7K MR-to-text instances spanning nine dialogue acts in the video game domain, with an average text length of around 14 words.  
Both E2E and ViGGO are closed-domain datasets.  
WikiTableText~\citep{bao2018table} is an open-domain D2T dataset consisting of approximately 13K table-to-text pairs extracted from Wikipedia tables.  
DART~\citep{nan2021dart} contains nearly 70K knowledge graph triplets, with an average text length of 34 words.  
WebNLG~\citep{gardent2017webnlg} focuses on RDF-to-text generation, comprising around 38K samples with an average text length of 30 words.  
Both DART and WebNLG are open-domain datasets.

\begin{table}[h]
\resizebox{\linewidth}{!}{
\begin{tabular}{@{}ccc@{}}
\toprule
family & model counts & parameters of each models \\ \midrule
Pythia & 8 & 70M, 160M, 410M, 1B, 1.4B, 2.8B, 6.9B, 12B \\
OPT & 6 & 130M, 350M, 1.3B, 2.7B, 6.7B, 13B \\
BLOOM & 5 & 0.56M, 1.1B, 1.7B, 3B, 7B \\ \bottomrule
\end{tabular}}
\caption{The three LLM families along with their models and corresponding sizes (M=million, B=billion).}
\label{tab:llm_model_families}
\end{table}

\subsection{Models}
We incorporate three widely used LLM families in our experiments: Pythia, OPT, and BLOOM.  
Examining multiple families offers broader insights than focusing on a single family.  
Pythia is a suite of eight decoder-only autoregressive models (70M–12B parameters), following a GPT-style~\citep{brown2020language} architecture with flash attention.
All Pythia models are trained on the Pile dataset in the same order.  
We consider OPT~\citep{zhang2022opt}, which includes six models, each being a decoder-only transformer (130M–13B parameters), trained on datasets including Reddit, the Pile, and RoBERTa, following the training details outlined in~\citep{brown2020language}.  
BLOOM~\citep{scao2022bloom} is another decoder-only LLM trained on the ROOT dataset.
We include six BLOOM models in our study.  
A summary of these LLM families, their models, and corresponding sizes is provided in~\autoref{tab:llm_model_families}.

\begin{figure*}[h]
    \centering
    \includegraphics[width=\textwidth]{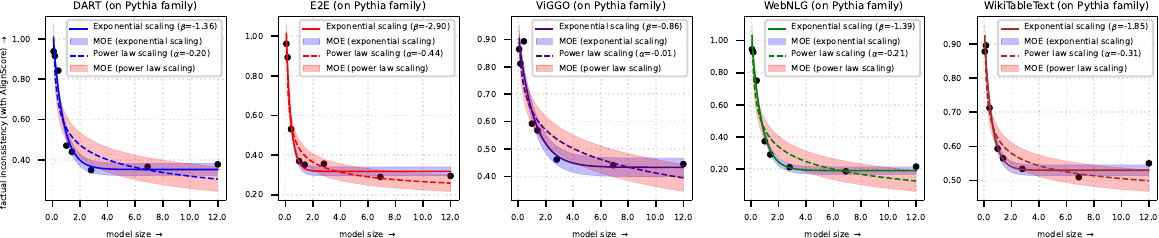}\\
    \vspace{2mm}
    \includegraphics[width=\textwidth]{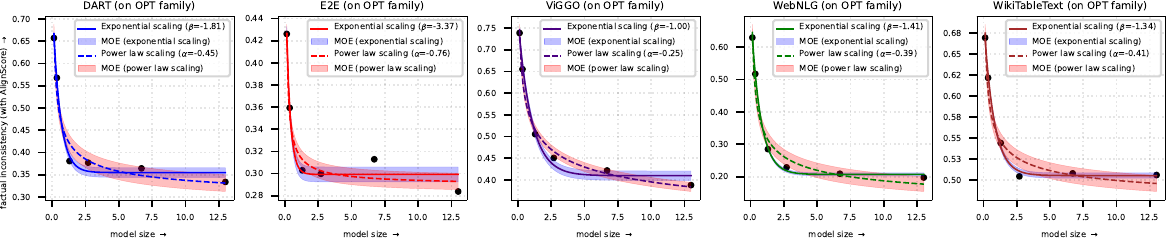}\\
    \vspace{2mm}
    \includegraphics[width=\textwidth]{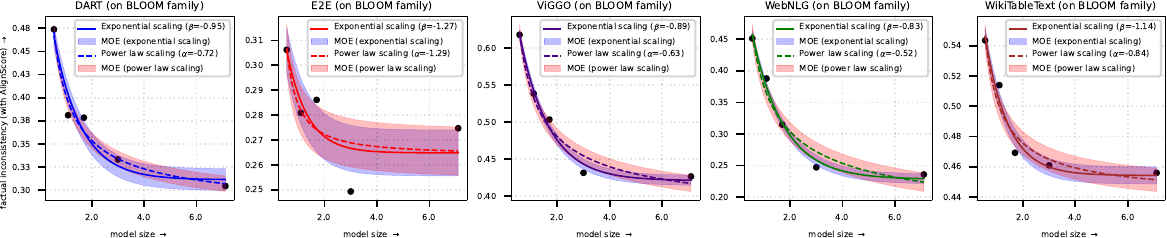}
    \caption{Visualization of exponential and power law scaling of factual inconsistency (\textsc{AlignScore}) across datasets and LLM families, with margin of error (MOE) and $95\%$ confidence intervals on residuals.}
    \label{fig:fit_alignscore}
\end{figure*}

\begin{table*}[h]
\resizebox{\textwidth}{!}{
\begin{tabular}{@{}cc|ccccc|ccccc@{}}
\toprule
\multicolumn{2}{c|}{} & \multicolumn{5}{c|}{Results of stage I} & \multicolumn{5}{c}{Results of stage II and III} \\ \midrule
LLM family & Scaling law & DART & E2E & ViGGO & WebNLG & WikiTableText & DART & E2E & ViGGO & WebNLG & WikiTableText \\ \midrule
 & Exponential & ${6.56}\text{e-}{04}$ & ${2.49}\text{e-}{03}$ & ${2.13}\text{e-}{04}$ & ${1.33}\text{e-}{03}$ & ${4.42}\text{e-}{03}$ & \cellcolor[HTML]{E67C73}\xmark & \cellcolor[HTML]{E67C73}\xmark & \cmark (\flower) & \cmark & \cmark \\
\multirow{-2}{*}{BLOOM} & Power law & ${8.42}\text{e-}{04}$ & ${3.07}\text{e-}{04}$ & ${2.44}\text{e-}{03}$ & ${1.30}\text{e-}{02}$ & ${2.50}\text{e-}{02}$ & \cmark & \cellcolor[HTML]{E67C73}\xmark & \cmark & \cellcolor[HTML]{E67C73}\xmark & \cellcolor[HTML]{E67C73}\xmark \\ \midrule
 & Exponential & ${4.15}\text{e-}{04}$ & ${9.15}\text{e-}{04}$ & ${4.16}\text{e-}{04}$ & ${1.30}\text{e-}{04}$ & \cellcolor[HTML]{E67C73}${2.82}\text{e+}{03}$ & \cmark (\flower) & \cmark (\flower) & \cmark & \cmark (\flower) & \cmark (\flower) \\
\multirow{-2}{*}{OPT} & Power law & ${2.27}\text{e-}{03}$ & ${1.38}\text{e-}{03}$ & ${1.94}\text{e-}{03}$ & ${2.08}\text{e-}{02}$ & \cellcolor[HTML]{E67C73}${6.37}\text{e+}{01}$ & \cmark & \cmark & \cellcolor[HTML]{FFD666}\cmark (\flower) & \cmark & \cmark \\ \midrule
 & Exponential & ${1.89}\text{e-}{03}$ & ${4.33}\text{e-}{02}$ & ${2.20}\text{e-}{03}$ & ${1.86}\text{e-}{03}$ & ${2.70}\text{e-}{03}$ & \cmark (\flower) & \cmark (\flower) & \cmark (\flower) & \cmark (\flower) & \cmark (\flower) \\
\multirow{-2}{*}{Pythia} & Power law & ${1.47}\text{e-}{02}$ & ${2.71}\text{e-}{01}$ & ${1.58}\text{e-}{02}$ & ${5.74}\text{e-}{03}$ & ${1.83}\text{e-}{02}$ & \cmark & \cmark & \cmark & \cmark & \cmark \\ \bottomrule
\end{tabular}}
\caption{Results of the validation framework (all three stages) for both scaling laws of factual inconsistency (\textsc{AlignScore}). High held-out losses (Stage I) are highlighted in red. \cmark/\xmark~indicates pass/fail (also marked in red) in the goodness-of-fit test (Stage II), while \flower~denotes the effective scaling law from Stage III.}
\label{tab:nucleus_alignscore}
\vspace{-3.5mm}
\end{table*}

\subsection{Fine-tuning for D2T}
All LLMs are fine-tuned separately on each of the five D2T datasets. Given the large model sizes, full fine-tuning is computationally expensive.
To mitigate this, we use QLoRA (Quantized Low-Rank Adapter)~\citep{dettmers2023qlora}, a parameter-efficient fine-tuning method, with a learning rate of $1.00\text{e-}{04}$ and $r=16$ (reduced rank) for the attention module.

\subsection{Quantification for Factual Inconsistency}
We define factual inconsistency as the inverse of factual consistency, computed as \(1 - z\) (where \(z\) is the factual consistency score ranging from 0 to 1).
To evaluate factual inconsistency in LLMs for D2T, we use four state-of-the-art automatic metrics that strongly correlate with human judgments: \textsc{AlignScore} (measures consistency through information alignment)~\citep{zha2023alignscore}, \textsc{QAFactEval} (assesses consistency via question generation and answering)~\citep{fabbri2022qafacteval}, \textsc{SummaC-conv} (leverages natural language inference)~\citep{laban2022summac}, and \textsc{UniEval-fact} (employs unified training)~\citep{zhong2022towards}.
Given their high agreement with human annotations, these metrics provide a strong foundation for our study.

\subsection{Decoding Strategies}
Given the importance of decoding strategies in D2T, we include both deterministic (greedy and beam search) and stochastic (nucleus and top-k sampling) methods for a comprehensive analysis. 
However, due to space constraints, we present here the results using nucleus sampling, while results for other strategies are provided in the appendix (\autoref{sec:appendix}).

\section{Results}
This section presents our empirical results and the validation framework's evaluation of factual inconsistency scaling in D2T based on the two scaling laws.
We report findings from the standpoint of automatic metrics used to assess factual inconsistency.

\begin{figure*}[h]
    \centering
    \includegraphics[width=\textwidth]{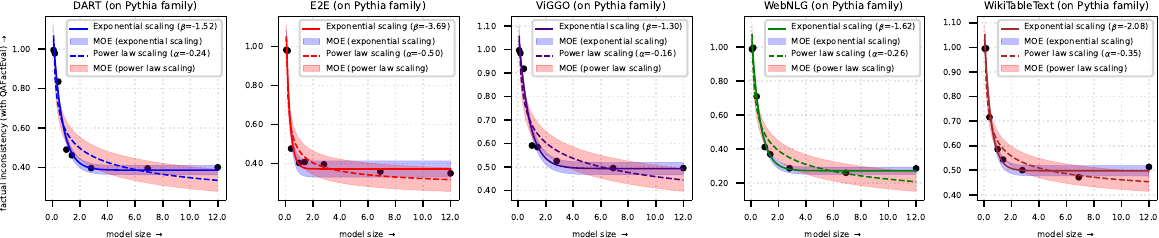}\\
    \vspace{2mm}
    \includegraphics[width=\textwidth]{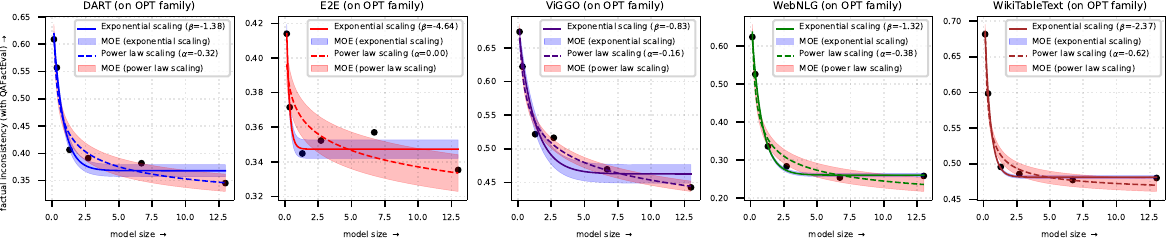}\\
    \vspace{2mm}
    \includegraphics[width=\textwidth]{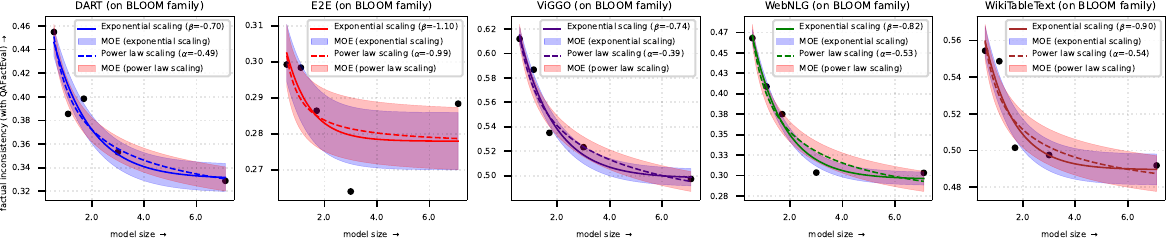}
    \caption{Visualization of exponential and power law scaling of factual inconsistency (\textsc{QAFactEval}) across datasets and LLM families, with margin of error (MOE) and $95\%$ confidence intervals on residuals.}
    \label{fig:fit_qafacteval}
\end{figure*}

\begin{table*}[h]
\resizebox{\textwidth}{!}{
\begin{tabular}{@{}cc|ccccc|ccccc@{}}
\toprule
\multicolumn{2}{c|}{} & \multicolumn{5}{c|}{Results of stage I} & \multicolumn{5}{c}{Results of stage II and III} \\ \midrule
LLM family & Scaling law & DART & E2E & ViGGO & WebNLG & WikiTableText & DART & E2E & ViGGO & WebNLG & WikiTableText \\ \midrule
 & Exponential & ${6.26}\text{e-}{04}$ & ${1.73}\text{e-}{03}$ & ${3.10}\text{e-}{04}$ & ${6.15}\text{e-}{04}$ & ${1.59}\text{e-}{02}$ & \cellcolor[HTML]{E67C73}\xmark & \cellcolor[HTML]{E67C73}\xmark & \cmark (\flower) & \cmark & \cellcolor[HTML]{E67C73}\xmark \\
\multirow{-2}{*}{BLOOM} & Power law & ${4.11}\text{e-}{04}$ & ${4.59}\text{e-}{03}$ & ${2.04}\text{e-}{03}$ & ${1.87}\text{e-}{03}$ & ${3.23}\text{e-}{01}$ & \cellcolor[HTML]{E67C73}\xmark & \cellcolor[HTML]{E67C73}\xmark & \cmark & \cellcolor[HTML]{E67C73}\xmark & \cellcolor[HTML]{E67C73}\xmark \\ \midrule
 & Exponential & ${3.57}\text{e-}{03}$ & \cellcolor[HTML]{E67C73}${1.28}\text{e+}{01}$ & ${6.56}\text{e-}{04}$ & ${2.12}\text{e-}{03}$ & ${1.67}\text{e-}{05}$ & \cmark (\flower) & \cmark & \cmark (\flower) & \cmark (\flower) & \cmark (\flower) \\
\multirow{-2}{*}{OPT} & Power law & ${1.41}\text{e-}{02}$ & ${2.81}\text{e-}{04}$ & ${4.43}\text{e-}{04}$ & ${8.45}\text{e-}{03}$ & ${5.99}\text{e-}{03}$ & \cmark & \cellcolor[HTML]{E67C73}\xmark & \cmark & \cmark & \cmark \\ \midrule
 & Exponential & ${1.89}\text{e-}{03}$ & ${4.33}\text{e-}{02}$ & ${2.20}\text{e-}{03}$ & ${1.86}\text{e-}{03}$ & ${2.70}\text{e-}{03}$ & \cmark (\flower) & \cmark (\flower) & \cmark (\flower) & \cmark (\flower) & \cmark (\flower) \\
\multirow{-2}{*}{Pythia} & Power law & ${1.47}\text{e-}{02}$ & ${2.71}\text{e-}{01}$ & ${1.58}\text{e-}{02}$ & ${5.74}\text{e-}{03}$ & ${1.83}\text{e-}{02}$ & \cmark & \cmark & \cmark & \cmark & \cmark \\ \bottomrule
\end{tabular}}
\caption{Results of the validation framework (all three stages) for exponential and power law scaling of factual inconsistency (\textsc{QAFactEval}). High held-out losses (Stage I) are highlighted in red. \cmark/\xmark~indicates pass/fail (also marked in red) in the goodness-of-fit test (Stage II), while \flower~denotes the effective scaling law from Stage III.}
\label{tab:nucleus_qafacteval}
\vspace{-3.5mm}
\end{table*}

\begin{figure*}[h]
    \centering
    \includegraphics[width=\textwidth]{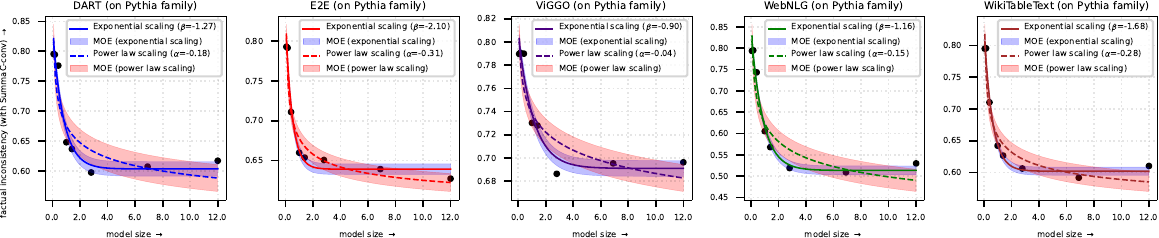}\\
    \vspace{2mm}
    \includegraphics[width=\textwidth]{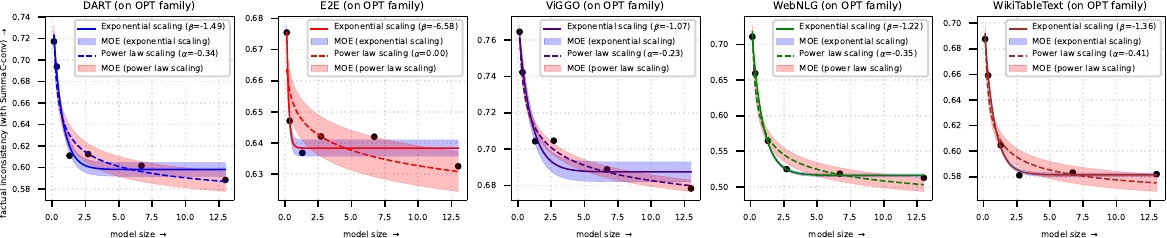}\\
    \vspace{2mm}
    \includegraphics[width=\textwidth]{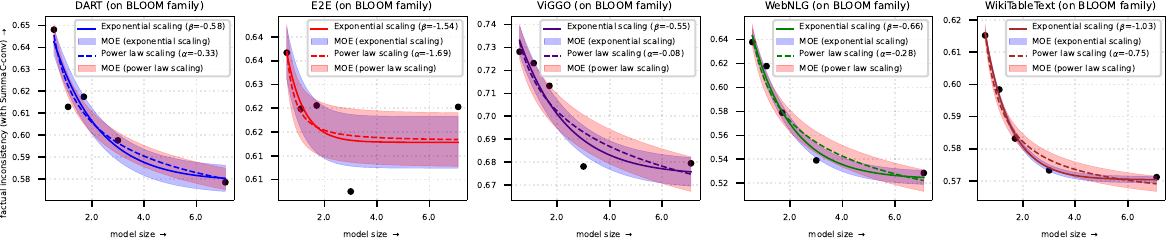}
    \caption{Visualization of exponential and power law scaling of factual inconsistency (\textsc{SummaC-conv}) across datasets and LLM families, with margin of error (MOE) and $95\%$ confidence intervals on residuals.}
    \label{fig:fit_summac}
\end{figure*}

\begin{table*}[h]
\resizebox{\textwidth}{!}{
\begin{tabular}{@{}cc|ccccc|ccccc@{}}
\toprule
\multicolumn{2}{c|}{} & \multicolumn{5}{c|}{Results of stage I} & \multicolumn{5}{c}{Results of stage II and III} \\ \midrule
LLM family & Scaling law & DART & E2E & ViGGO & WebNLG & WikiTableText & DART & E2E & ViGGO & WebNLG & WikiTableText \\ \midrule
 & Exponential & ${1.70}\text{e-}{04}$ & ${4.25}\text{e-}{04}$ & ${7.60}\text{e-}{04}$ & ${1.00}\text{e-}{03}$ & ${1.94}\text{e-}{05}$ & \cellcolor[HTML]{E67C73}\xmark & \cellcolor[HTML]{E67C73}\xmark & \cellcolor[HTML]{E67C73}\xmark & \cmark & \cmark (\flower) \\
\multirow{-2}{*}{BLOOM} & Power law & ${7.00}\text{e-}{05}$ & ${1.28}\text{e-}{04}$ & ${4.27}\text{e-}{04}$ & ${1.77}\text{e-}{03}$ & ${9.60}\text{e-}{04}$ & \cellcolor[HTML]{E67C73}\xmark & \cellcolor[HTML]{E67C73}\xmark & \cellcolor[HTML]{E67C73}\xmark & \cellcolor[HTML]{E67C73}\xmark & \cmark \\ \midrule
 & Exponential & ${1.71}\text{e-}{04}$ & ${9.87}\text{e-}{05}$ & ${8.72}\text{e-}{05}$ & ${2.83}\text{e-}{05}$ & \cellcolor[HTML]{E67C73}${1.01}\text{e+}{03}$ & \cmark (\flower) & \cmark (\flower) & \cmark (\flower) & \cmark (\flower) & \cmark (\flower) \\
\multirow{-2}{*}{OPT} & Power law & ${1.60}\text{e-}{03}$ & ${1.36}\text{e-}{04}$ & ${9.33}\text{e-}{05}$ & ${5.39}\text{e-}{03}$ & \cellcolor[HTML]{E67C73}${5.48}\text{e+}{19}$ & \cmark & \cmark & \cmark & \cmark & \cmark \\ \midrule
 & Exponential & ${4.17}\text{e-}{04}$ & ${3.36}\text{e-}{04}$ & ${5.00}\text{e-}{04}$ & ${3.66}\text{e-}{04}$ & ${1.18}\text{e-}{04}$ & \cmark (\flower) & \cmark (\flower) & \cmark (\flower) & \cmark (\flower) & \cmark (\flower) \\
\multirow{-2}{*}{Pythia} & Power law & ${1.43}\text{e-}{03}$ & ${1.79}\text{e-}{04}$ & ${1.40}\text{e-}{02}$ & ${2.23}\text{e-}{03}$ & ${2.19}\text{e-}{02}$ & \cmark & \cmark & \cmark & \cmark & \cmark \\ \bottomrule
\end{tabular}}
\caption{Results of the validation framework (all three stages) for exponential and power law scaling of factual inconsistency (\textsc{SummaC-conv}). High held-out losses (Stage I) are highlighted in red. \cmark/\xmark~indicates pass/fail (also marked in red) in the goodness-of-fit test (Stage II), while \flower~denotes the effective scaling law from Stage III.}
\label{tab:nucleus_summac}
\vspace{-3.5mm}
\end{table*}

\begin{figure*}[h]
    \centering
    \includegraphics[width=\textwidth]{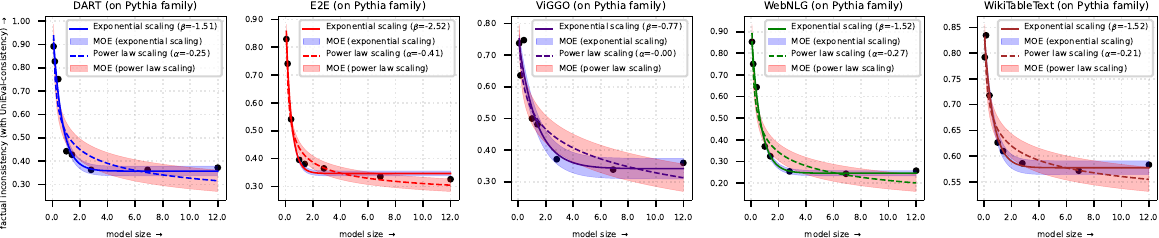}\\
    \vspace{2mm}
    \includegraphics[width=\textwidth]{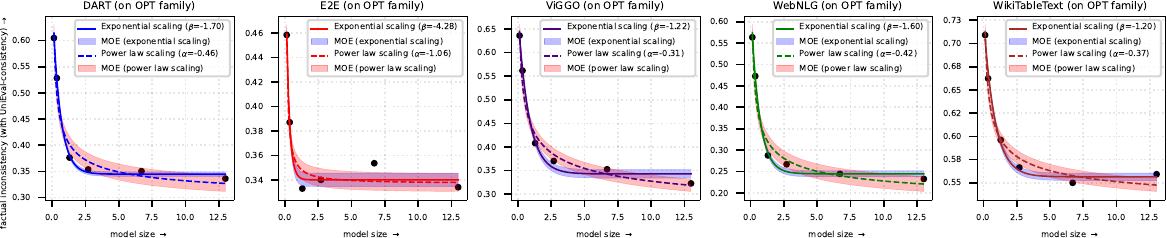}\\
    \vspace{2mm}
    \includegraphics[width=\textwidth]{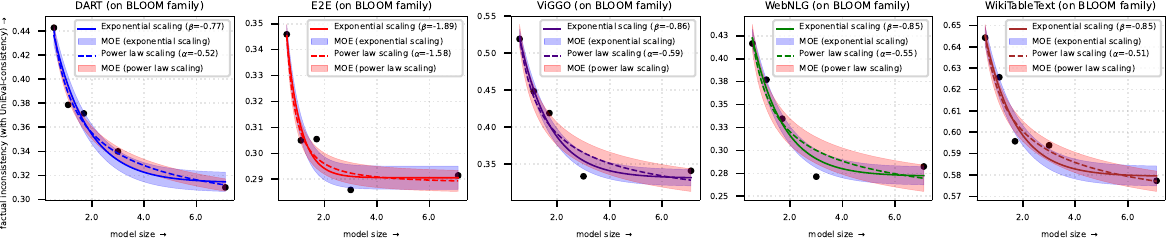}
    \caption{Visualization of exponential and power law scaling of factual inconsistency (\textsc{UniEval-fact}) across datasets and LLM families, with margin of error (MOE) and $95\%$ confidence intervals on residuals.}
    \label{fig:fit_unieval}
\end{figure*}

\begin{table*}[h]
\resizebox{\textwidth}{!}{
\begin{tabular}{@{}cc|ccccc|ccccc@{}}
\toprule
\multicolumn{2}{c|}{} & \multicolumn{5}{c|}{Results of stage I} & \multicolumn{5}{c}{Results of stage II and III} \\ \midrule
LLM family & Scaling law & DART & E2E & ViGGO & WebNLG & WikiTableText & DART & E2E & ViGGO & WebNLG & WikiTableText \\ \midrule
 & Exponential & 2.87E-04 & 1.16E-04 & 2.01E-04 & 3.66E-03 & 2.13E-04 & \cmark (\flower) & \cellcolor[HTML]{E67C73}\xmark & \cmark & \cellcolor[HTML]{E67C73}\xmark & \cmark \\
\multirow{-2}{*}{BLOOM} & Power law & 3.71E-04 & 6.11E-05 & 1.57E-03 & 8.66E-03 & 6.57E-04 & \cmark & \cellcolor[HTML]{E67C73}\xmark & \cellcolor[HTML]{E67C73}\xmark & \cellcolor[HTML]{E67C73}\xmark & \cellcolor[HTML]{FFD666}\cmark (\flower) \\ \midrule
 & Exponential & 7.78E-05 & 6.45E-04 & 3.32E-03 & 1.62E-04 & 8.35E-05 & \cmark (\flower) & \cmark (\flower) & \cmark (\flower) & \cmark (\flower) & \cmark (\flower) \\
\multirow{-2}{*}{OPT} & Power law & 6.25E-03 & 5.70E-04 & 3.93E-03 & 5.47E-03 & 3.18E-03 & \cmark & \cmark & \cmark & \cmark & \cmark \\ \midrule
 & Exponential & 6.34E-04 & 2.63E-04 & 2.08E-03 & 4.58E-04 & 4.54E-05 & \cmark (\flower) & \cmark (\flower) & \cmark (\flower) & \cmark (\flower) & \cmark (\flower) \\
\multirow{-2}{*}{Pythia} & Power law & 7.29E-03 & 8.28E-03 & 1.83E-01 & 1.13E-02 & 4.19E-03 & \cmark & \cmark & \cmark & \cmark & \cmark \\ \bottomrule
\end{tabular}}
\caption{Results of the validation framework (all three stages) for exponential and power law scaling of factual inconsistency (\textsc{UniEval-fact}). High held-out losses (Stage I) are highlighted in red. \cmark/\xmark~indicates pass/fail (also marked in red) in the goodness-of-fit test (Stage II), while \flower~denotes the effective scaling law from Stage III.}
\label{tab:nucleus_unieval}
\vspace{-3.5mm}
\end{table*}

\subsection{Findings from \textsc{AlignScore}}
\autoref{fig:fit_alignscore} illustrates both fitted scaling laws for factual inconsistency measured by \textsc{AlignScore}.
Exponential scaling generally outperforms power law scaling, except for minor deviations, such as larger margins of error (MOE) in the BLOOM family for the E2E dataset.  
\autoref{tab:nucleus_alignscore} presents our statistical validation results.
Note that we successfully verified the normality assumption on residuals before applying our validation framework.
Most Huber loss values are low, confirming the predictive reliability of both scaling laws. 
However, in several cases within the BLOOM family, one or both scaling models fail the goodness-of-fit test, while both pass for OPT and Pythia.  
This confirms that reliable predictive performance alone is not always sufficient to pass the goodness-of-fit test.
Stage III results (\autoref{tab:nucleus_alignscore}) indicate that exponential scaling is generally preferred over power law scaling, except for the ViGGO dataset with OPT (marked in yellow).
The stringent significance level of the final test further strengthens our conclusion, providing compelling evidence that factual inconsistency in D2T, when measured using \textsc{AlignScore}, consistently follows an exponential scaling pattern with respect to LLM size, rather than a power law trend.

\subsection{Findings from \textsc{QAFactEval}}
\autoref{fig:fit_qafacteval} suggests that when factual inconsistency is measured using \textsc{QAFactEval}, both scaling laws fit well across most datasets and LLM families.
Power law scaling shows a larger margin of error (MOE) compared to exponential scaling, with extremely high MOE for OPT and BLOOM families on the E2E dataset.
Here we also verified the normality assumption before applying our validation framework.
Low losses in stage I results (\autoref{tab:nucleus_qafacteval}) indicate strong predictive performance for both scaling laws.
Stage II and III results (\autoref{tab:nucleus_qafacteval}) show that both scaling law not qualify for goodness-of-fit in the BLOOM family, while both laws are qualified for goodness-of-fit in Pythia and OPT, with exponential scaling outperforming power law scaling.
Thus, based on Pythia and OPT, we observe that exponential scaling appears more suitable when factual inconsistency is measured using \textsc{QAFactEval} in D2T.

\subsection{Findings from \textsc{SummaC-conv}}
\autoref{fig:fit_summac} shows both fitted scaling laws when factual inconsistency is measured using \textsc{SummaC-conv}.
In most cases, exponential scaling provides a better fit than power law scaling.
Some datasets, like E2E, exhibit a slightly larger margin of error (MOE), particularly in the BLOOM model family, which can be considered a minor exception. 
From~\autoref{tab:nucleus_summac}, we observe that, except for a few cases in WikiTableText, the Huber loss values remain low, ensuring the predictive quality of both scaling laws. 
Additionally, both scaling law fails to qualify in several goodness-of-fit tests across multiple datasets in BLOOOM family.
However, in the other two LLM families, where both scaling laws pass the test, exponential scaling consistently outperforms power law scaling.

\subsection{Findings from \textsc{UniEval-fact}}
\autoref{fig:fit_unieval} illustrates how both scaling laws perform across all LLM families and D2T datasets when measuring factual inconsistency with \textsc{UniEval-fact}. 
\autoref{tab:nucleus_unieval} presents the validation framework results, consistently showing that exponential scaling captures factual consistency better than power law scaling.
The only exception is WikiTableText in the BLOOM family (highlighted in yellow), where power law scaling surpasses exponential scaling.

\section{Discussion}
Our results demonstrate that when factual inconsistency is measured using the four automatic metrics, exponential scaling consistently outperforms power law scaling in most cases.
While a few exceptions arise, particularly within the BLOOM family, we consider these anomalies to be outliers, likely due to the limited number of models in the family (only five).
Beyond this, both scaling laws exhibit minimal margins of error across all plots, reinforcing their predictive reliability.
The acceptance of exponential scaling for factual inconsistency in D2T relative to LLM size suggests a rapid initial decline in factual inconsistency up to approximately 3–4 billion parameters, after which it stabilizes.
Furthermore, in cases where exponential scaling does not show high margins of error, the scaling rate ($\beta$) consistently falls within the range of $-1.8$ to $-0.6$.
Understanding this range of exponential scaling rates could be crucial for predicting the performance of LLMs in D2T tasks concerning factual inconsistency, providing valuable insights for future model development and evaluation.

\section{Conclusion}
This paper shows that factual inconsistency in D2T generally follows exponential scaling with respect to LLM size, rather than the commonly assumed power law scaling.
Our findings are validated through a structured three-stage statistical framework, ensuring robustness in our conclusions.
Moreover, we conduct a comprehensive empirical study using three major LLM families across five D2T datasets, measuring factual inconsistency inversely with four state-of-the-art consistency metrics.
We believe these findings will help researchers and practitioners select appropriate model sizes to achieve specific levels of factual consistency.
All results in this paper are based on a parameter-efficient fine-tuning approach (QLoRA) for LLMs.
However, in-context learning and prompting strategies are not considered in this study, which we leave as future work.

\section{Limitations}
While our study provides a thorough empirical analysis of scaling laws for factual consistency in LLMs, validated through a structured three-stage framework, it is important to acknowledge the following limitations:   
\begin{enumerate}
    \item \textbf{Empirical basis without theoretical guarantee.} Our findings are entirely based on empirical observations, relying on the datasets and LLM families incorporated in this study. We do not provide a formal theoretical guarantee for the observed scaling behavior, making our conclusions inherently dependent on the data and models used.
    \item \textbf{Non-universality of scaling law parameters.} Scaling law parameters are not universally applicable across different datasets, models, and task domains.
    While our results indicate a strong preference for exponential scaling, this does not guarantee that the same trend will persist across all datasets or model architectures, even when using the same set of parameters.
    Therefore, applying these scaling laws— including our own findings—requires careful consideration and validation within the specific context of use.
    \item \textbf{Reliance on automated metrics without human evaluation.} In this study, factual inconsistency is estimated inversely from automated factual consistency metrics. 
    While these metrics have demonstrated strong correlations with human judgments, we do not incorporate direct human evaluations of factual consistency.
    This remains a limitation of our work, though it presents a clear direction for future research to further validate and refine our findings with human annotation studies.
\end{enumerate}

\section*{Acknowledgments}
This research is partially supported by the Indo-French Centre for the Promotion of Advanced Research (IFCPAR/CEFIPRA) through CSRP Project No. 6702-2.

\bibliography{references/scaling_and_stats,references/final_base,references/statistical_test}

\appendix

\begin{figure*}[h]
    \centering
    \includegraphics[width=\textwidth]{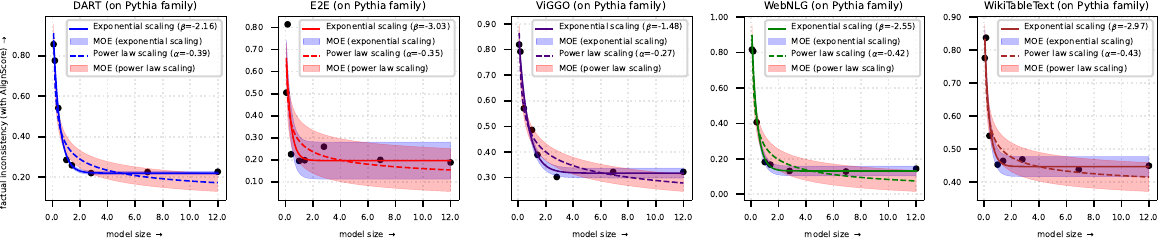}\\
    \vspace{2mm}
    \includegraphics[width=\textwidth]{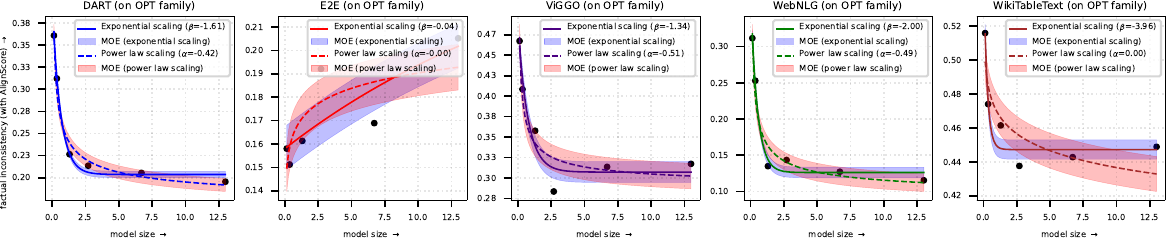}\\
    \vspace{2mm}
    \includegraphics[width=\textwidth]{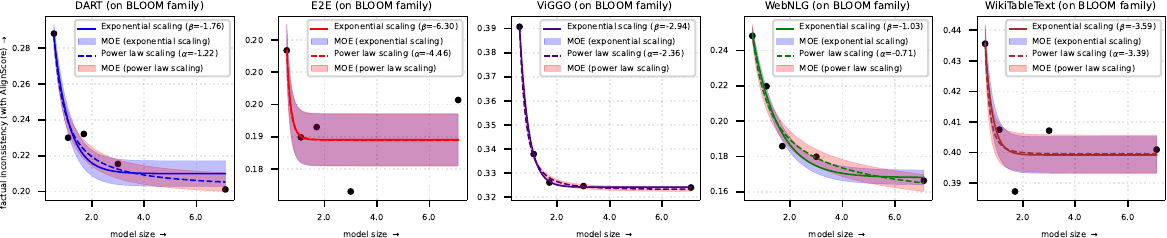}
    \caption{Visualization of exponential and power law scaling of factual inconsistency (\textsc{AlignScore}) across datasets and LLM families, with the margin of error (MOE) and $95\%$ confidence intervals on residuals. Texts generated using the \ul{greedy search decoding algorithm}. Aberrant behavior of the E2E dataset with the OPT family, as discussed in~\ref{subsec:critical}.}
    \label{fig:fit_alignscore_greedy}
\end{figure*}

\begin{table*}[h]
\resizebox{\textwidth}{!}{
\begin{tabular}{@{}cc|ccccc|ccccc@{}}
\toprule
\multicolumn{2}{c|}{} & \multicolumn{5}{c|}{Results of stage I} & \multicolumn{5}{c}{Results of stage II and III} \\ \midrule
LLM family & Scaling law & DART & E2E & ViGGO & WebNLG & WikiTableText & DART & E2E & ViGGO & WebNLG & WikiTableText \\ \midrule
 & Exponential & ${4.36}\text{e-}{04}$ & ${5.32}\text{e-}{06}$ & ${5.41}\text{e-}{05}$ & ${2.69}\text{e-}{04}$ & \cellcolor[HTML]{E67C73}${4.75}\text{e+}{02}$ & \cellcolor[HTML]{E67C73}\xmark & \cellcolor[HTML]{E67C73}\xmark & \cmark (\flower) & \cmark (\flower) & \cellcolor[HTML]{E67C73}\xmark \\
\multirow{-2}{*}{BLOOM} & Power law & ${2.75}\text{e-}{04}$ & ${4.85}\text{e-}{05}$ & ${2.19}\text{e-}{06}$ & ${2.88}\text{e-}{05}$ & ${3.30}\text{e-}{04}$ & \cmark & \cellcolor[HTML]{E67C73}\xmark & \cmark & \cmark & \cellcolor[HTML]{E67C73}\xmark \\ \midrule
 & Exponential & ${3.12}\text{e-}{05}$ & \cellcolor[HTML]{E67C73}${1.15}\text{e+}{02}$ & ${7.46}\text{e-}{03}$ & ${2.37}\text{e-}{01}$ & ${2.00}\text{e-}{04}$ & \cmark (\flower) & \cellcolor[HTML]{E67C73}\xmark & \cmark (\flower) & \cmark (\flower) & \cmark \\
\multirow{-2}{*}{OPT} & Power law & ${2.17}\text{e-}{03}$ & ${1.11}\text{e-}{04}$ & \cellcolor[HTML]{E67C73}${7.62}\text{e+}{19}$ & ${6.96}\text{e-}{03}$ & ${1.87}\text{e-}{04}$ & \cmark & \cellcolor[HTML]{E67C73}\xmark & \cmark & \cmark & \xmark \\ \midrule
 & Exponential & ${2.60}\text{e-}{04}$ & ${1.96}\text{e-}{02}$ & ${1.05}\text{e-}{03}$ & ${6.05}\text{e-}{03}$ & \cellcolor[HTML]{E67C73}${1.80}\text{e+}{01}$ & \cmark (\flower) & \cmark & \cmark (\flower) & \cmark (\flower) & \cmark (\flower) \\
\multirow{-2}{*}{Pythia} & Power law & ${2.22}\text{e-}{02}$ & \cellcolor[HTML]{E67C73}${3.19}\text{e+}{00}$ & ${2.02}\text{e-}{03}$ & ${1.37}\text{e-}{01}$ & ${1.15}\text{e-}{01}$ & \cmark & \cellcolor[HTML]{E67C73}\xmark & \cmark & \cmark & \cmark \\ \bottomrule
\end{tabular}}
\caption{\ul{[Case: greedy decoding]} Results of the validation framework (all three stages) for exponential and power law scaling of factual inconsistency (\textsc{AlignScore}). High held-out losses (Stage I) are highlighted in red. \cmark/\xmark~indicates pass/fail (also marked in red) in the goodness-of-fit test (Stage II), while \flower~denotes the effective scaling law from Stage III.}
\label{tab:greedy_alignscore}
\end{table*}

\section{Appendix}
\label{sec:appendix}
In the main paper, we discussed the crucial role of decoding strategies in data-to-text generation (D2T) and presented results based on nucleus sampling.
Here, we extend our analysis by presenting empirical results from our validation framework for both power law and exponential scaling, using three additional decoding strategies—greedy, beam search, and top-k decoding.
Among these, greedy and beam search are deterministic, while top-k decoding falls under stochastic methods.
For our experiments, we use beam search with a beam size of 3 and top-k decoding with $k = 640$ (sample size).

\subsection{Discussion}
\label{subsec:critical}
\begin{figure*}[h]
    \centering
    \includegraphics[width=\textwidth]{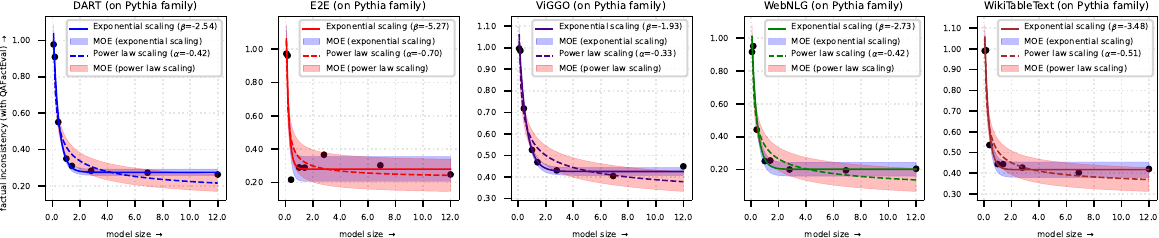}\\
    \vspace{2mm}
    \includegraphics[width=\textwidth]{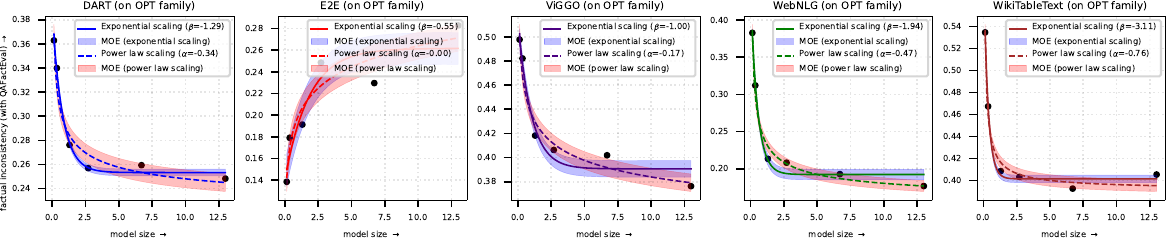}\\
    \vspace{2mm}
    \includegraphics[width=\textwidth]{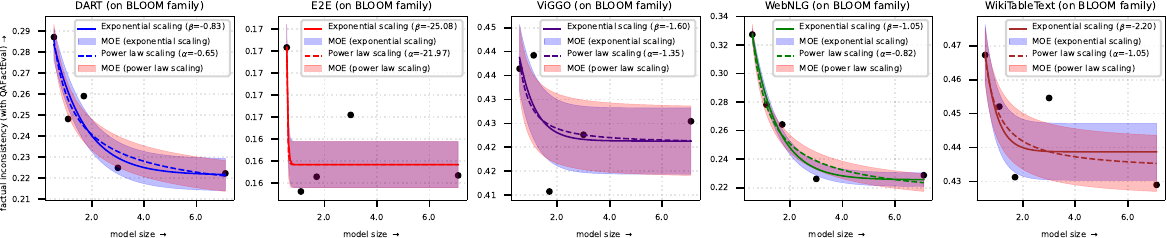}
    \caption{Visualization of exponential and power law scaling of factual inconsistency (\textsc{QAFactEval}) across datasets and LLM families, with margin of error (MOE) and $95\%$ confidence intervals on residuals. Texts generated using the \ul{greedy search decoding algorithm}. Aberrant behavior of the E2E dataset with the OPT family, as discussed in~\ref{subsec:critical}.}
    \label{fig:fit_qafacteval_greedy}
\end{figure*}

\begin{figure*}[h]
    \centering
    \includegraphics[width=\textwidth]{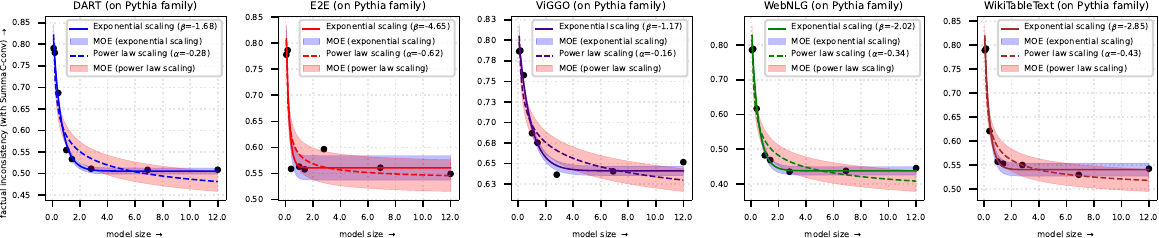}\\
    \vspace{2mm}
    \includegraphics[width=\textwidth]{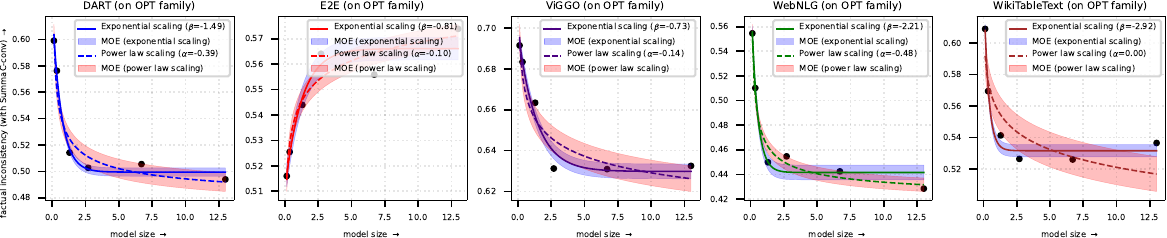}\\
    \vspace{2mm}
    \includegraphics[width=\textwidth]{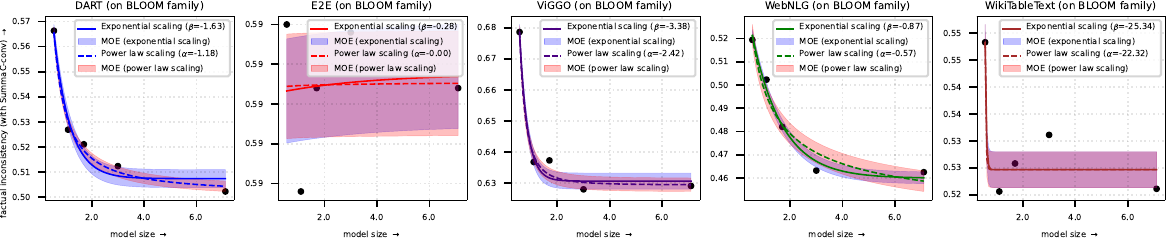}
    \caption{Visualization of exponential and power law scaling of factual inconsistency (\textsc{SummaC-conv}) across datasets and LLM families, with margin of error (MOE) and $95\%$ confidence intervals on residuals. Texts generated using the \ul{greedy search decoding algorithm}. The unusual behavior of the E2E dataset with the OPT and BLOOM families is discussed in~\ref{subsec:critical}.}
    \label{fig:fit_summac_greedy}
\end{figure*}

\begin{figure*}[h]
    \centering
    \includegraphics[width=\textwidth]{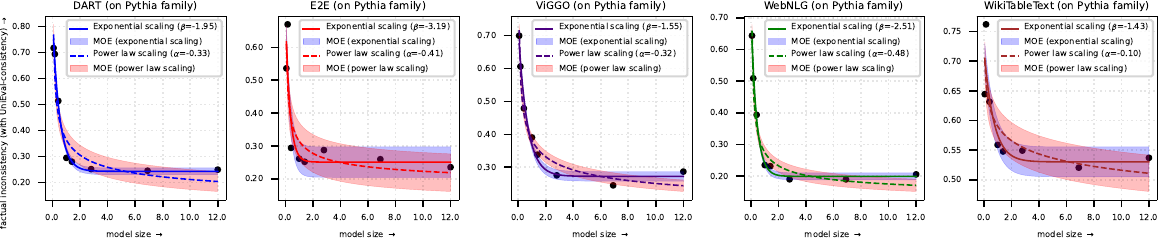}\\
    \vspace{2mm}
    \includegraphics[width=\textwidth]{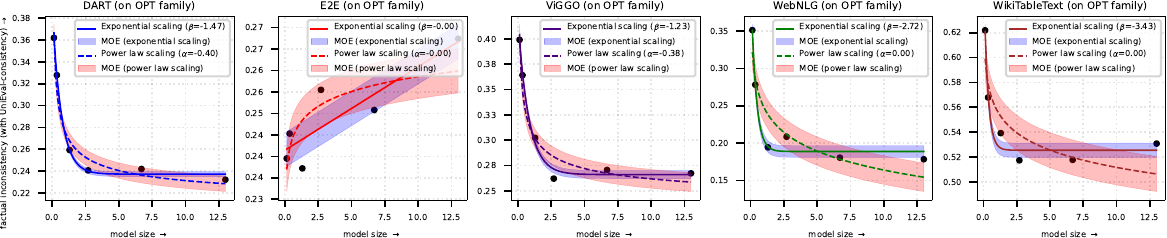}\\
    \vspace{2mm}
    \includegraphics[width=\textwidth]{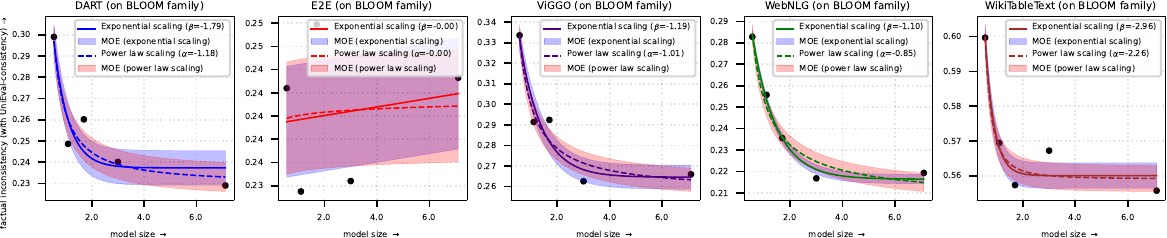}
    \caption{Visualization of exponential and power law scaling of factual inconsistency (\textsc{UniEval-fact}) across datasets and LLM families, with margin of error (MOE) and $95\%$ confidence intervals on residuals. Texts generated using the \ul{greedy search decoding algorithm}.
    The unusual behavior of the E2E dataset with the OPT family is discussed in~\ref{subsec:critical}.}
    \label{fig:fit_unieval_greedy}
\end{figure*}

\begin{figure*}[h]
    \centering
    \includegraphics[width=\textwidth]{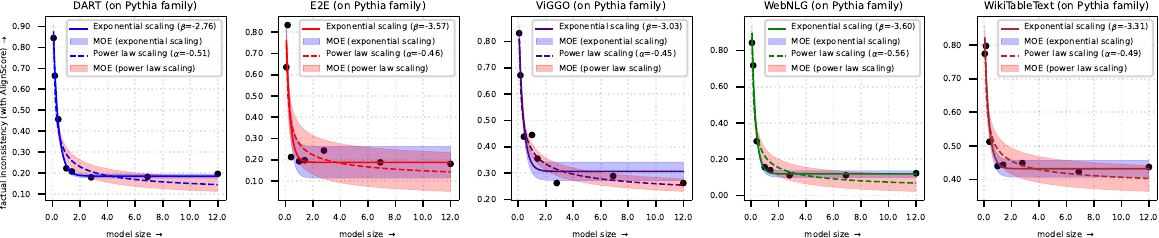}\\
    \vspace{2mm}
    \includegraphics[width=\textwidth]{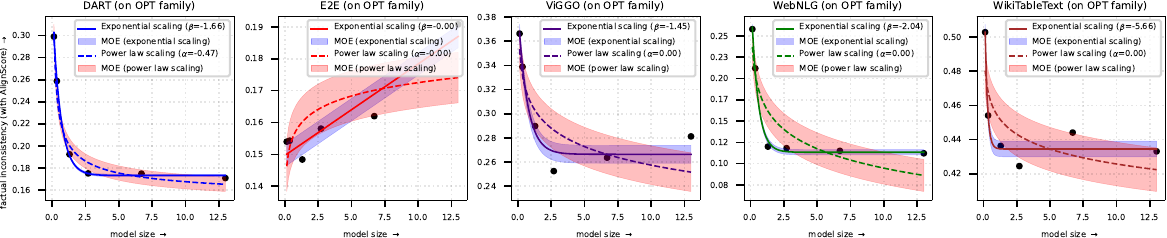}\\
    \vspace{2mm}
    \includegraphics[width=\textwidth]{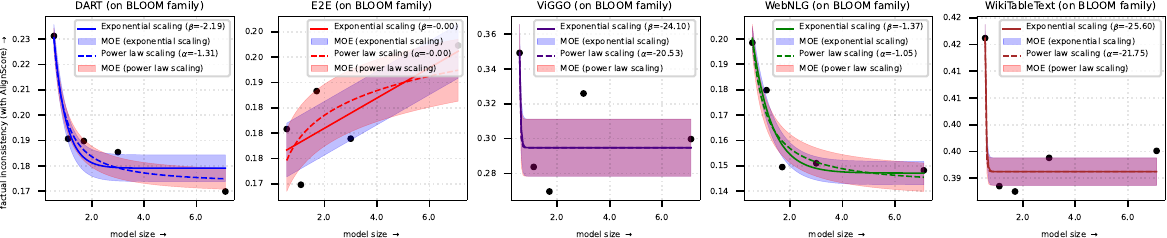}
    \caption{Visualization of exponential and power law scaling of factual inconsistency (\textsc{AlignScore}) across datasets and LLM families, with the margin of error (MOE) and $95\%$ confidence intervals on residuals. Texts generated using the \ul{beam search decoding algorithm}.
    The unusual behavior of the E2E dataset with the OPT and BLOOM families is discussed in~\ref{subsec:critical}.}
    \label{fig:fit_alignscore_beam}
\end{figure*}

\begin{figure*}[h]
    \centering
    \includegraphics[width=\textwidth]{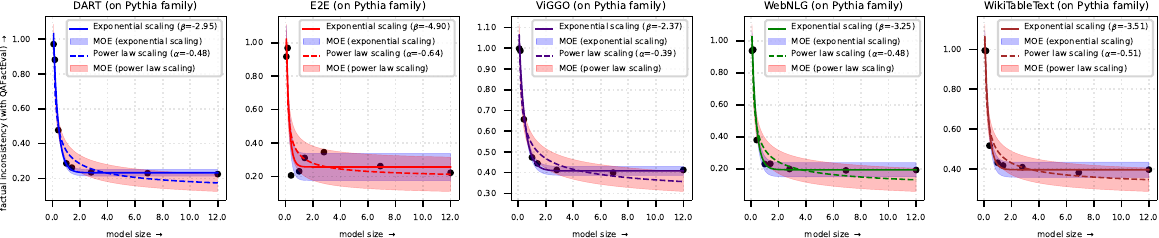}\\
    \vspace{2mm}
    \includegraphics[width=\textwidth]{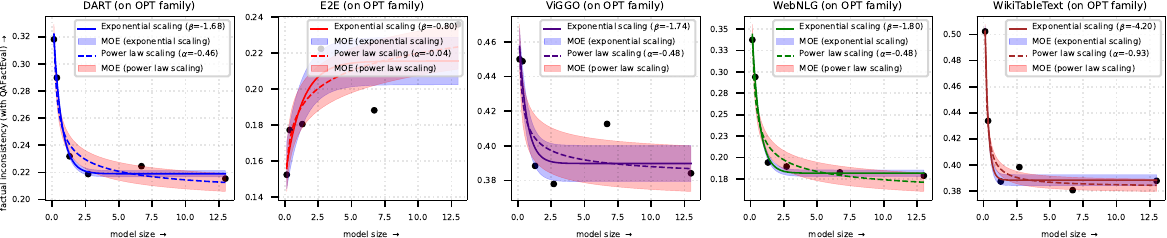}\\
    \vspace{2mm}
    \includegraphics[width=\textwidth]{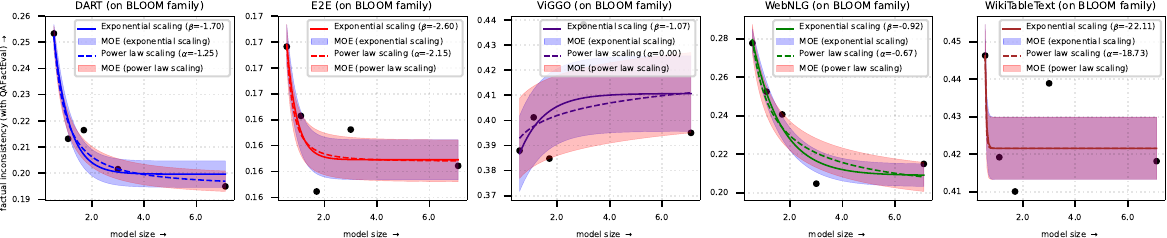}
    \caption{Visualization of exponential and power law scaling of factual inconsistency (\textsc{QAFactEval}) across datasets and LLM families, with margin of error (MOE) and $95\%$ confidence intervals on residuals. Texts generated using the \ul{beam search decoding algorithm}.
    The unusual behavior of the E2E and ViGGO dataset with the OPT family is discussed in~\ref{subsec:critical}.}
    \label{fig:fit_qafacteval_beam}
\end{figure*}

\begin{figure*}[h]
    \centering
    \includegraphics[width=\textwidth]{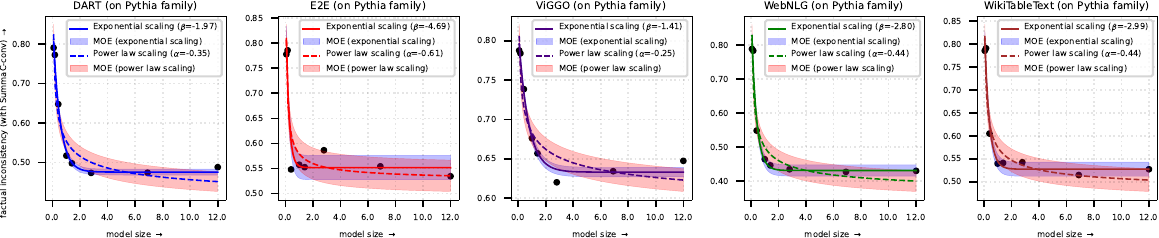}\\
    \vspace{2mm}
    \includegraphics[width=\textwidth]{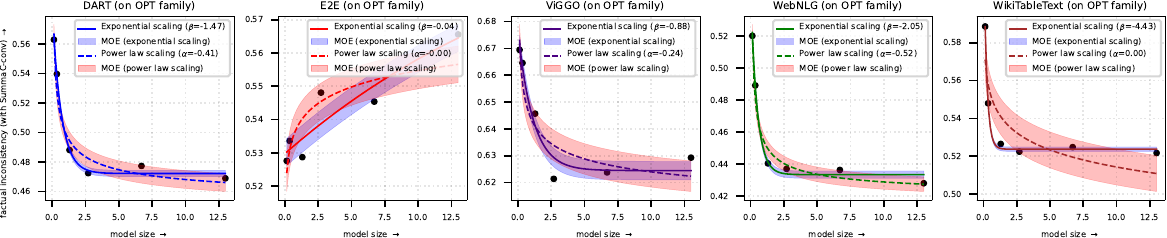}\\
    \vspace{2mm}
    \includegraphics[width=\textwidth]{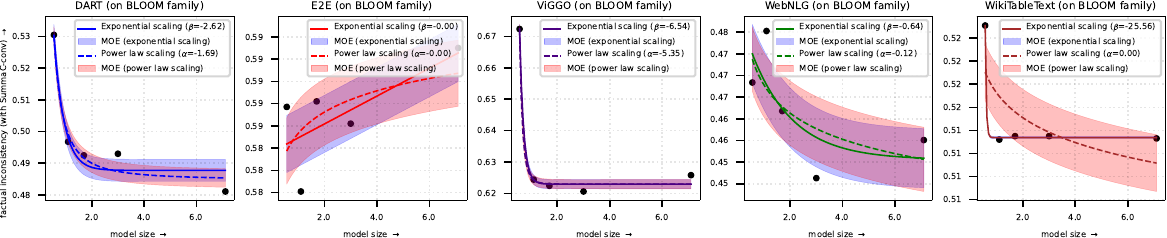}
    \caption{Visualization of exponential and power law scaling of factual inconsistency (\textsc{SummaC-conv}) across datasets and LLM families, with margin of error (MOE) and $95\%$ confidence intervals on residuals. Texts generated using the \ul{beam search decoding algorithm}.
    The unusual behavior of the E2E dataset with the OPT and BLLOM families is discussed in~\ref{subsec:critical}.}
    \label{fig:fit_summac_beam}
\end{figure*}

\begin{figure*}[h]
    \centering
    \includegraphics[width=\textwidth]{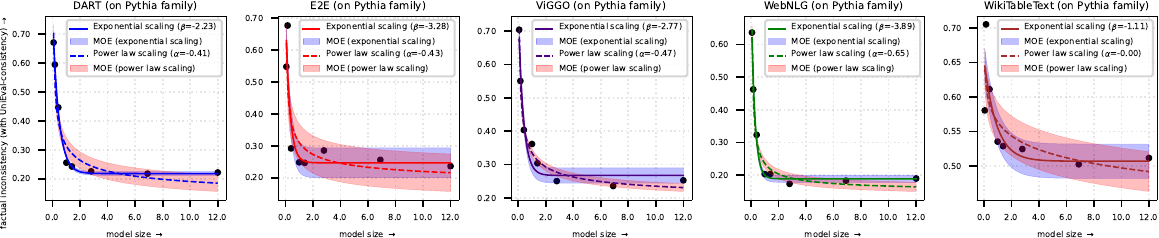}\\
    \vspace{2mm}
    \includegraphics[width=\textwidth]{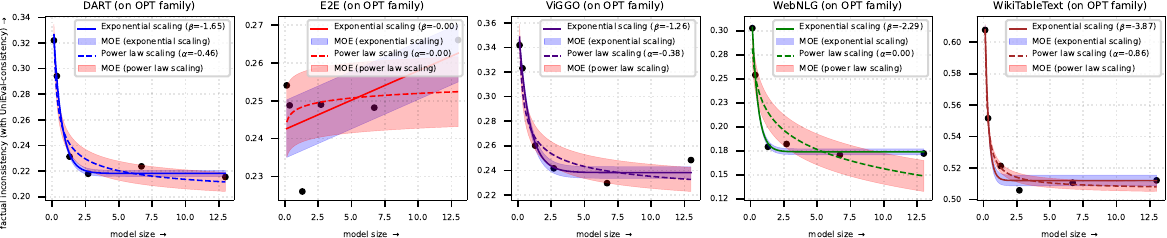}\\
    \vspace{2mm}
    \includegraphics[width=\textwidth]{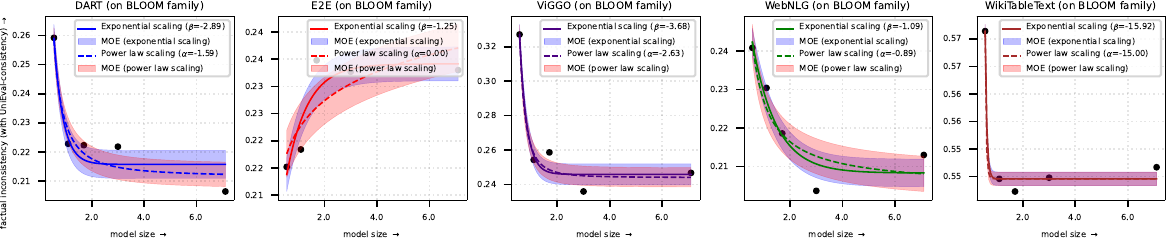}
    \caption{Visualization of exponential and power law scaling of factual inconsistency (\textsc{UniEval-fact}) across datasets and LLM families, with margin of error (MOE) and $95\%$ confidence intervals on residuals. Texts generated using the \ul{beam search decoding algorithm}.
    The unusual behavior of the E2E dataset with the OPT and BLOOM families is discussed in~\ref{subsec:critical}.}
    \label{fig:fit_unieval_beam}
\end{figure*}

Across all three decoding strategies, we consistently observe that \textbf{exponential scaling outperforms power law scaling} in nearly all cases for the Pythia and OPT LLM families across the four factual inconsistency metrics.
While this trend is dominant, we identify a few noteworthy exceptions:

\begin{enumerate}
    \item \textbf{Goodness-of-fit test failures in BLOOM and OPT.} In the BLOOM and OPT families, we frequently find cases where the scaling laws fail to qualify the goodness-of-fit test (Stage II), despite demonstrating low predictive loss in Stage I.
    This indicates that strong predictive performance alone is not always sufficient for a model to align well with the expected scaling trend.
    In other words, higher predictive performance does not necessarily imply goodness-of-fit.
    
    \item \textbf{High margin of error in E2E dataset.} The margin of error (with a $95\%$ confidence interval) tends to be significantly higher in the E2E dataset, particularly for the BLOOM and OPT model families.
    This suggests a higher variance in factual inconsistency measurements, potentially due to dataset-specific characteristics or the way these models generalize.
    
    \item \textbf{Aberrant behavior in E2E and ViGGO with deterministic decoding.} A particularly intriguing anomaly is observed in the E2E and ViGGO datasets (\Cref{fig:fit_alignscore_greedy,fig:fit_qafacteval_greedy,fig:fit_summac_greedy,fig:fit_unieval_greedy,fig:fit_alignscore_beam,fig:fit_qafacteval_beam,fig:fit_summac_beam,fig:fit_unieval_beam}), where factual inconsistency increases with LLM model size under deterministic decoding strategies (greedy search and beam search). 
    This contradicts the general trend seen with stochastic decoding strategies (nucleus and top-k sampling), where inconsistency decreases with model size. 
    We hypothesize that this aberrant behavior may be attributed to one or both of the following factors:
    \begin{itemize}
        \item \textbf{Deterministic decoding bias.} Since greedy search and beam search select high-likelihood tokens, they might reinforce factual errors present in the training data rather than mitigating them.
        \item \textbf{Closed-domain nature of E2E and ViGGO.} These datasets focus on highly structured, domain-specific content, which may lead to overfitting in larger models when using deterministic decoding.
    \end{itemize}
\end{enumerate}

\begin{table*}[h]
\resizebox{\textwidth}{!}{
\begin{tabular}{@{}cc|ccccc|ccccc@{}}
\toprule
\multicolumn{2}{c|}{} & \multicolumn{5}{c|}{Results of stage I} & \multicolumn{5}{c}{Results of stage II and III} \\ \midrule
LLM family & Scaling law & DART & E2E & ViGGO & WebNLG & WikiTableText & DART & E2E & ViGGO & WebNLG & WikiTableText \\ \midrule
 & Exponential & ${1.47}\text{e-}{04}$ & ${4.07}\text{e-}{05}$ & ${1.60}\text{e-}{04}$ & ${1.11}\text{e-}{04}$ & ${2.35}\text{e-}{04}$ & \cellcolor[HTML]{E67C73}\xmark & \cellcolor[HTML]{E67C73}\xmark & \cellcolor[HTML]{E67C73}\xmark & \cmark (\flower) & \xmark \\
\multirow{-2}{*}{BLOOM} & Power law & ${1.96}\text{e-}{04}$ & ${4.17}\text{e-}{05}$ & ${1.07}\text{e-}{04}$ & ${3.86}\text{e-}{04}$ & ${6.78}\text{e-}{05}$ & \cellcolor[HTML]{E67C73}\xmark & \cellcolor[HTML]{E67C73}\xmark & \cellcolor[HTML]{E67C73}\xmark & \cmark & \xmark \\ \midrule
 & Exponential & ${2.59}\text{e-}{05}$ & ${3.98}\text{e-}{03}$ & ${2.99}\text{e-}{04}$ & ${1.88}\text{e-}{03}$ & ${1.54}\text{e-}{04}$ & \cmark (\flower) & \cmark (\flower) & \cmark & \cmark & \cmark (\flower) \\
\multirow{-2}{*}{OPT} & Power law & ${1.90}\text{e-}{04}$ & ${4.07}\text{e-}{04}$ & ${1.48}\text{e-}{04}$ & ${3.44}\text{e-}{03}$ & ${5.28}\text{e-}{04}$ & \cmark & \cmark & \cellcolor[HTML]{FFD666}\cmark (\flower) & \cellcolor[HTML]{FFD666}\cmark (\flower) & \cmark \\ \midrule
 & Exponential & ${2.89}\text{e-}{03}$ & \cellcolor[HTML]{E67C73}${2.06}\text{e+}{01}$ & ${1.47}\text{e-}{03}$ & ${1.91}\text{e-}{02}$ & ${8.07}\text{e-}{03}$ & \cmark (\flower) & \cmark (\flower) & \cmark (\flower) & \cmark (\flower) & \cmark (\flower) \\
\multirow{-2}{*}{Pythia} & Power law & ${4.84}\text{e-}{02}$ & ${3.79}\text{e-}{02}$ & ${4.50}\text{e-}{02}$ & ${2.83}\text{e-}{01}$ & ${2.88}\text{e-}{01}$ & \cmark & \cmark & \cmark & \cmark & \cmark \\ \bottomrule
\end{tabular}}
\caption{\ul{[Case: greedy decoding]} Results of the validation framework (all three stages) for exponential and power law scaling of factual inconsistency (\textsc{QAFactEval}). High held-out losses (Stage I) are highlighted in red. \cmark/\xmark~indicates pass/fail (also marked in red) in the goodness-of-fit test (Stage II), while \flower~denotes the effective scaling law from Stage III.}
\label{tab:greedy_qafacteval}
\end{table*}

\begin{table*}[h]
\resizebox{\textwidth}{!}{
\begin{tabular}{@{}cc|ccccc|ccccc@{}}
\toprule
\multicolumn{2}{c|}{} & \multicolumn{5}{c|}{Results of stage I} & \multicolumn{5}{c}{Results of stage II and III} \\ \midrule
LLM family & Scaling law & DART & E2E & ViGGO & WebNLG & WikiTableText & DART & E2E & ViGGO & WebNLG & WikiTableText \\ \midrule
 & Exponential & ${4.46}\text{e-}{05}$ & ${6.49}\text{e-}{06}$ & ${2.80}\text{e-}{04}$ & ${1.49}\text{e-}{04}$ & ${2.24}\text{e-}{03}$ & \cmark (\flower) & \cellcolor[HTML]{E67C73}\xmark & \cmark (\flower) & \cmark & \cellcolor[HTML]{E67C73}\xmark \\
\multirow{-2}{*}{BLOOM} & Power law & ${2.44}\text{e-}{04}$ & ${8.29}\text{e-}{07}$ & ${1.57}\text{e-}{04}$ & ${2.83}\text{e-}{05}$ & ${1.41}\text{e-}{04}$ & \cmark & \cellcolor[HTML]{E67C73}\xmark & \cmark & \cellcolor[HTML]{E67C73}\xmark & \cellcolor[HTML]{E67C73}\xmark \\ \midrule
 & Exponential & ${1.14}\text{e-}{03}$ & ${1.81}\text{e-}{02}$ & ${6.09}\text{e-}{03}$ & ${6.51}\text{e-}{03}$ & ${5.25}\text{e-}{03}$ & \cmark (\flower) & \cmark & \cmark (\flower) & \cmark & \cmark (\flower) \\
\multirow{-2}{*}{OPT} & Power law & ${1.12}\text{e-}{02}$ & ${1.91}\text{e-}{02}$ & ${3.07}\text{e-}{02}$ & ${4.41}\text{e-}{03}$ & ${1.12}\text{e-}{01}$ & \cmark & \cellcolor[HTML]{FFD666}\cmark (\flower) & \cmark & \cellcolor[HTML]{FFD666}\cmark (\flower) & \cellcolor[HTML]{E67C73}\xmark \\ \midrule
 & Exponential & ${2.88}\text{e-}{04}$ & \cellcolor[HTML]{E67C73}${5.56}\text{e+}{00}$ & ${9.95}\text{e-}{05}$ & ${1.16}\text{e-}{03}$ & ${1.31}\text{e-}{03}$ & \cmark (\flower) & \cmark (\flower) & \cmark (\flower) & \cmark (\flower) & \cmark (\flower) \\
\multirow{-2}{*}{Pythia} & Power law & ${2.39}\text{e-}{01}$ & ${2.61}\text{e+}{03}$ & ${4.16}\text{e-}{04}$ & ${3.87}\text{e-}{02}$ & ${1.80}\text{e-}{03}$ & \cmark & \cmark & \cmark & \cmark & \cmark \\ \bottomrule
\end{tabular}}
\caption{\ul{[Case: greedy decoding]} Results of the validation framework (all three stages) for exponential and power law scaling of factual inconsistency (\textsc{SummaC-conv}). High held-out losses (Stage I) are highlighted in red. \cmark/\xmark~indicates pass/fail (also marked in red) in the goodness-of-fit test (Stage II), while \flower~denotes the effective scaling law from Stage III.}
\label{tab:greedy_summac}
\end{table*}

\begin{table*}[h]
\resizebox{\textwidth}{!}{
\begin{tabular}{@{}cc|ccccc|ccccc@{}}
\toprule
\multicolumn{2}{c|}{} & \multicolumn{5}{c|}{Results of stage I} & \multicolumn{5}{c}{Results of stage II and III} \\ \midrule
LLM family & Scaling law & DART & E2E & ViGGO & WebNLG & WikiTableText & DART & E2E & ViGGO & WebNLG & WikiTableText \\ \midrule
 & Exponential & ${6.76}\text{e-}{04}$ & ${4.41}\text{e-}{05}$ & ${7.84}\text{e-}{05}$ & ${6.20}\text{e-}{05}$ & ${5.08}\text{e-}{05}$ & \cellcolor[HTML]{E67C73}\xmark & \cellcolor[HTML]{E67C73}\xmark & \cellcolor[HTML]{E67C73}\xmark & \cmark (\flower) & \cellcolor[HTML]{E67C73}\xmark \\
\multirow{-2}{*}{BLOOM} & Power law & ${2.40}\text{e-}{04}$ & ${3.14}\text{e-}{05}$ & ${8.77}\text{e-}{05}$ & ${1.62}\text{e-}{03}$ & ${1.65}\text{e-}{04}$ & \cellcolor[HTML]{E67C73}\xmark & \cellcolor[HTML]{E67C73}\xmark & \cellcolor[HTML]{E67C73}\xmark & \cellcolor[HTML]{E67C73}\cmark & \cellcolor[HTML]{E67C73}\xmark \\ \midrule
 & Exponential & ${1.19}\text{e-}{05}$ & ${1.30}\text{e-}{04}$ & ${6.18}\text{e-}{05}$ & ${7.44}\text{e-}{04}$ & ${1.40}\text{e-}{04}$ & \cmark (\flower) & \cellcolor[HTML]{E67C73}\xmark & \cmark (\flower) & \cmark (\flower) & \cmark \\
\multirow{-2}{*}{OPT} & Power law & ${7.95}\text{e-}{04}$ & ${5.41}\text{e-}{05}$ & ${2.21}\text{e-}{04}$ & ${2.73}\text{e-}{04}$ & ${4.62}\text{e-}{04}$ & \cmark & \cellcolor[HTML]{E67C73}\xmark & \cmark & \cmark & \cellcolor[HTML]{E67C73}\xmark \\ \midrule
 & Exponential & ${1.35}\text{e-}{03}$ & ${4.14}\text{e-}{02}$ & ${6.82}\text{e-}{04}$ & ${9.60}\text{e-}{04}$ & ${5.77}\text{e-}{03}$ & \cmark (\flower) & \cmark (\flower) & \cmark (\flower) & \cmark (\flower) & \cmark (\flower) \\
\multirow{-2}{*}{Pythia} & Power law & ${3.32}\text{e-}{03}$ & ${6.94}\text{e-}{01}$ & ${4.18}\text{e-}{04}$ & ${8.26}\text{e-}{04}$ & ${1.08}\text{e-}{02}$ & \cmark & \cmark & \cmark & \cmark & \cmark \\ \bottomrule
\end{tabular}}
\caption{\ul{[Case: greedy decoding]} Results of the validation framework (all three stages) for exponential and power law scaling of factual inconsistency (\textsc{UniEval-fact}). High held-out losses (Stage I) are highlighted in red. \cmark/\xmark~indicates pass/fail (also marked in red) in the goodness-of-fit test (Stage II), while \flower~denotes the effective scaling law from Stage III.}
\label{tab:greedy_unieval}
\end{table*}

\begin{table*}[h]
\resizebox{\textwidth}{!}{
\begin{tabular}{@{}cc|ccccc|ccccc@{}}
\toprule
\multicolumn{2}{c|}{} & \multicolumn{5}{c|}{Results of stage I} & \multicolumn{5}{c}{Results of stage II and III} \\ \midrule
LLM family & Scaling law & DART & E2E & ViGGO & WebNLG & WikiTableText & DART & E2E & ViGGO & WebNLG & WikiTableText \\ \midrule
 & Exponential & ${2.30}\text{e-}{04}$ & ${4.46}\text{e-}{05}$ & ${1.20}\text{e-}{03}$ & \cellcolor[HTML]{FFCCC9}${2.35}\text{e+}{02}$ & ${1.05}\text{e-}{04}$ & \cellcolor[HTML]{FFCCC9}\xmark & \cellcolor[HTML]{FFCCC9}\xmark & \cellcolor[HTML]{FFCCC9}\xmark & \cellcolor[HTML]{FFCCC9}\xmark & \cellcolor[HTML]{FFCCC9}\xmark \\
\multirow{-2}{*}{BLOOM} & Power law & ${1.69}\text{e-}{04}$ & ${9.06}\text{e-}{05}$ & ${9.50}\text{e-}{04}$ & ${5.79}\text{e-}{05}$ & ${1.94}\text{e-}{04}$ & \cmark & \cellcolor[HTML]{FFCCC9}\xmark & \cellcolor[HTML]{FFCCC9}\xmark & \cellcolor[HTML]{FFCCC9}\xmark & \cellcolor[HTML]{FFCCC9}\xmark \\ \midrule
 & Exponential & ${5.75}\text{e-}{06}$ & ${6.66}\text{e-}{05}$ & \cellcolor[HTML]{FFCCC9}${7.95}\text{e+}{02}$ & ${1.57}\text{e-}{03}$ & ${8.62}\text{e-}{03}$ & \cmark (\flower) & \cmark & \cmark & \cmark & \cmark \\
\multirow{-2}{*}{OPT} & Power law & ${4.15}\text{e-}{04}$ & ${1.20}\text{e-}{04}$ & ${1.47}\text{e-}{03}$ & ${3.25}\text{e-}{02}$ & ${4.00}\text{e-}{04}$ & \cmark & \cellcolor[HTML]{FFCCC9}\xmark & \cellcolor[HTML]{FFCCC9}\xmark & \cellcolor[HTML]{FFCCC9}\xmark & \cellcolor[HTML]{FFCCC9}\xmark \\ \midrule
 & Exponential & ${2.67}\text{e-}{03}$ & ${2.74}\text{e-}{01}$ & ${4.12}\text{e-}{03}$ & ${1.48}\text{e-}{02}$ & ${1.00}\text{e-}{02}$ & \cmark (\flower) & \cmark & \cmark (\flower) & \cmark (\flower) & \cmark (\flower) \\
\multirow{-2}{*}{Pythia} & Power law & \cellcolor[HTML]{FFCCC9}${1.79}\text{e+}{00}$ & \cellcolor[HTML]{FFCCC9}${3.62}\text{e+}{00}$ & ${3.52}\text{e-}{03}$ & ${6.17}\text{e-}{02}$ & ${1.90}\text{e-}{03}$ & \cmark & \cellcolor[HTML]{FFCCC9}\xmark & \cmark & \cmark & \cmark \\ \bottomrule
\end{tabular}}
\caption{\ul{[Case: beam search decoding]} Results of the validation framework (all three stages) for exponential and power law scaling of factual inconsistency (\textsc{AlignScore}). High held-out losses (Stage I) are highlighted in red. \cmark/\xmark~indicates pass/fail (also marked in red) in the goodness-of-fit test (Stage II), while \flower~denotes the effective scaling law from Stage III.}
\label{tab:beam_alignscore}
\end{table*}

\begin{table*}[h]
\resizebox{\textwidth}{!}{
\begin{tabular}{@{}cc|rrrrr|lccll@{}}
\toprule
\multicolumn{2}{c|}{} & \multicolumn{5}{c|}{Results of stage I} & \multicolumn{5}{c}{Results of stage II and III} \\ \midrule
LLM family & Scaling law & \multicolumn{1}{c}{DART} & \multicolumn{1}{c}{E2E} & \multicolumn{1}{c}{ViGGO} & \multicolumn{1}{c}{WebNLG} & \multicolumn{1}{c|}{WikiTableText} & \multicolumn{1}{c}{DART} & E2E & ViGGO & \multicolumn{1}{c}{WebNLG} & \multicolumn{1}{c}{WikiTableText} \\ \midrule
 & Exponential & ${1.91}\text{e-}{04}$ & ${5.37}\text{e-}{06}$ & ${2.93}\text{e-}{04}$ & ${9.19}\text{e-}{05}$ & ${2.14}\text{e-}{04}$ & \multicolumn{1}{c}{\cellcolor[HTML]{E67C73}\xmark} & \cellcolor[HTML]{E67C73}\xmark & \cellcolor[HTML]{E67C73}\xmark & \multicolumn{1}{c}{\cellcolor[HTML]{E67C73}\xmark} & \multicolumn{1}{c}{\cellcolor[HTML]{E67C73}\xmark} \\
\multirow{-2}{*}{BLOOM} & Power law & ${1.37}\text{e-}{04}$ & ${8.13}\text{e-}{06}$ & ${4.54}\text{e-}{04}$ & ${1.52}\text{e-}{03}$ & ${9.32}\text{e-}{05}$ & \multicolumn{1}{c}{\cellcolor[HTML]{E67C73}\xmark} & \cellcolor[HTML]{E67C73}\xmark & \cellcolor[HTML]{E67C73}\xmark & \multicolumn{1}{c}{\cellcolor[HTML]{E67C73}\xmark} & \multicolumn{1}{c}{\cellcolor[HTML]{E67C73}\xmark} \\ \midrule
 & Exponential & \multicolumn{1}{c}{\cellcolor[HTML]{E67C73}${1.39}\text{e+}{02}$} & \multicolumn{1}{c}{\cellcolor[HTML]{E67C73}${2.02}\text{e+}{02}$} & ${1.40}\text{e-}{03}$ & ${2.24}\text{e-}{04}$ & ${7.31}\text{e-}{02}$ & \cmark (\flower) & \cellcolor[HTML]{E67C73}\xmark & \cellcolor[HTML]{E67C73}\xmark & \cmark (\flower) & \cmark (\flower) \\
\multirow{-2}{*}{OPT} & Power law & ${1.09}\text{e-}{03}$ & ${6.17}\text{e-}{03}$ & ${7.27}\text{e-}{03}$ & ${7.60}\text{e-}{03}$ & \multicolumn{1}{c|}{\cellcolor[HTML]{E67C73}${1.25}\text{e+}{02}$} & \cmark & \cellcolor[HTML]{E67C73}\xmark & \cellcolor[HTML]{E67C73}\xmark & \cmark & \cmark \\ \midrule
 & Exponential & ${2.39}\text{e-}{03}$ & ${2.10}\text{e+}{01}$ & ${2.87}\text{e-}{04}$ & ${2.11}\text{e-}{02}$ & ${1.51}\text{e-}{02}$ & \cmark (\flower) & \multicolumn{1}{l}{\cmark} & \multicolumn{1}{l}{\cmark (\flower)} & \cmark (\flower) & \cmark (\flower) \\
\multirow{-2}{*}{Pythia} & Power law & ${3.79}\text{e-}{01}$ & ${3.64}\text{e+}{04}$ & ${9.37}\text{e-}{02}$ & ${1.87}\text{e-}{01}$ & ${1.46}\text{e-}{01}$ & \cmark & \cellcolor[HTML]{E67C73}\xmark & \multicolumn{1}{l}{\cmark} & \cmark & \cmark \\ \bottomrule
\end{tabular}}
\caption{\ul{[Case: beam search decoding]} Results of the validation framework (all three stages) for exponential and power law scaling of factual inconsistency (\textsc{QAFactEval}). High held-out losses (Stage I) are highlighted in red. \cmark/\xmark~indicates pass/fail (also marked in red) in the goodness-of-fit test (Stage II), while \flower~denotes the effective scaling law from Stage III.}
\label{tab:beam_qafacteval}
\end{table*}

\begin{table*}[h]
\resizebox{\textwidth}{!}{
\begin{tabular}{@{}cc|ccccc|ccccc@{}}
\toprule
\multicolumn{2}{c|}{} & \multicolumn{5}{c|}{Results of stage I} & \multicolumn{5}{c}{Results of stage II and III} \\ \midrule
LLM family & Scaling law & DART & E2E & ViGGO & WebNLG & WikiTableText & DART & E2E & ViGGO & WebNLG & WikiTableText \\ \midrule
 & Exponential & ${1.55}\text{e-}{04}$ & ${2.03}\text{e-}{05}$ & \cellcolor[HTML]{E67C73}${1.53}\text{e+}{01}$ & ${3.51}\text{e-}{04}$ & ${3.38}\text{e-}{01}$ & \cellcolor[HTML]{E67C73}\xmark & \cellcolor[HTML]{E67C73}\xmark & \cmark (\flower) & \cellcolor[HTML]{E67C73}\xmark & \cmark \\
\multirow{-2}{*}{BLOOM} & Power law & ${2.65}\text{e-}{05}$ & ${1.08}\text{e-}{04}$ & ${6.76}\text{e-}{04}$ & ${5.33}\text{e-}{05}$ & ${1.49}\text{e-}{05}$ & \cmark & \cellcolor[HTML]{E67C73}\xmark & \cmark & \cellcolor[HTML]{E67C73}\xmark & \cellcolor[HTML]{E67C73}\xmark \\ \midrule
 & Exponential & \cellcolor[HTML]{E67C73}${5.36}\text{e+}{01}$ & ${2.00}\text{e-}{04}$ & ${1.62}\text{e-}{04}$ & ${3.13}\text{e-}{05}$ & \cellcolor[HTML]{E67C73}${2.67}\text{e+}{01}$ & \cmark (\flower) & \cmark & \cmark & \cmark (\flower) & \cmark \\
\multirow{-2}{*}{OPT} & Power law & ${5.59}\text{e-}{04}$ & ${5.69}\text{e-}{05}$ & ${3.19}\text{e-}{03}$ & ${6.87}\text{e-}{05}$ & ${1.64}\text{e-}{04}$ & \cmark & \cellcolor[HTML]{E67C73}\xmark & \cellcolor[HTML]{E67C73}\xmark & \cmark & \cellcolor[HTML]{E67C73}\xmark \\ \midrule
 & Exponential & ${1.11}\text{e-}{04}$ & \cellcolor[HTML]{E67C73}${1.20}\text{e+}{01}$ & ${1.34}\text{e-}{04}$ & ${7.35}\text{e-}{04}$ & ${1.36}\text{e-}{03}$ & \cmark (\flower) & \cmark (\flower) & \cmark (\flower) & \cmark (\flower) & \cmark (\flower) \\
\multirow{-2}{*}{Pythia} & Power law & ${1.23}\text{e-}{03}$ & \cellcolor[HTML]{E67C73}${1.10}\text{e+}{04}$ & ${9.79}\text{e-}{04}$ & ${1.91}\text{e-}{02}$ & ${7.29}\text{e-}{04}$ & \cmark & \cmark & \cmark & \cmark & \cmark \\ \bottomrule
\end{tabular}}
\caption{\ul{[Case: beam search decoding]} Results of the validation framework (all three stages) for exponential and power law scaling of factual inconsistency (\textsc{SummaC-conv}). High held-out losses (Stage I) are highlighted in red. \cmark/\xmark~indicates pass/fail (also marked in red) in the goodness-of-fit test (Stage II), while \flower~denotes the effective scaling law from Stage III.}
\label{tab:beam_summac}
\end{table*}

\begin{table*}[h]
\resizebox{\textwidth}{!}{
\begin{tabular}{@{}cc|ccccc|ccccc@{}}
\toprule
\multicolumn{2}{c|}{} & \multicolumn{5}{c|}{Results of stage I} & \multicolumn{5}{c}{Results of stage II and III} \\ \midrule
LLM family & Scaling law & DART & E2E & ViGGO & WebNLG & WikiTableText & DART & E2E & ViGGO & WebNLG & WikiTableText \\ \midrule
 & Exponential & ${6.54}\text{e-}{05}$ & \cellcolor[HTML]{E67C73}${4.84}\text{e+}{02}$ & ${7.41}\text{e-}{04}$ & ${2.27}\text{e-}{04}$ & ${8.25}\text{e-}{05}$ & \cellcolor[HTML]{E67C73}\xmark & \cellcolor[HTML]{E67C73}\xmark & \cellcolor[HTML]{E67C73}\xmark & \cellcolor[HTML]{E67C73}\xmark & \cmark \\
\multirow{-2}{*}{BLOOM} & Power law & ${5.75}\text{e-}{05}$ & ${3.88}\text{e-}{05}$ & ${5.41}\text{e-}{05}$ & ${3.64}\text{e-}{05}$ & ${2.83}\text{e-}{06}$ & \cellcolor[HTML]{E67C73}\xmark & \cellcolor[HTML]{E67C73}\xmark & \cellcolor[HTML]{E67C73}\xmark & \cellcolor[HTML]{E67C73}\xmark & \cellcolor[HTML]{FFD666}\cmark (\flower) \\ \midrule
 & Exponential & \cellcolor[HTML]{E67C73}${9.29}\text{e+}{01}$ & ${6.84}\text{e-}{05}$ & ${5.15}\text{e-}{05}$ & ${1.05}\text{e-}{03}$ & ${1.82}\text{e-}{04}$ & \cmark (\flower) & \cellcolor[HTML]{E67C73}\xmark & \cmark (\flower) & \cmark & \cmark (\flower) \\
\multirow{-2}{*}{OPT} & Power law & ${4.10}\text{e-}{03}$ & ${1.15}\text{e-}{04}$ & \cellcolor[HTML]{E67C73}${1.91}\text{e+}{01}$ & ${7.27}\text{e-}{03}$ & ${3.50}\text{e-}{05}$ & \cmark & \cellcolor[HTML]{E67C73}\xmark & \cmark & \cellcolor[HTML]{E67C73}\xmark & \cmark \\ \midrule
 & Exponential & ${7.15}\text{e-}{05}$ & ${3.72}\text{e-}{02}$ & ${2.53}\text{e-}{03}$ & ${9.19}\text{e-}{04}$ & ${3.21}\text{e-}{03}$ & \cmark (\flower) & \cmark (\flower) & \cmark (\flower) & \cmark (\flower) & \cmark \\
\multirow{-2}{*}{Pythia} & Power law & ${9.70}\text{e-}{01}$ & \cellcolor[HTML]{E67C73}${4.10}\text{e+}{01}$ & ${2.56}\text{e-}{04}$ & ${5.47}\text{e-}{01}$ & ${5.11}\text{e-}{03}$ & \cmark & \cmark & \cmark & \cmark & \cellcolor[HTML]{E67C73}\xmark \\ \bottomrule
\end{tabular}}
\caption{\ul{[Case: beam search decoding]} Results of the validation framework (all three stages) for exponential and power law scaling of factual inconsistency (\textsc{UniEval-fact}). High held-out losses (Stage I) are highlighted in red. \cmark/\xmark~indicates pass/fail (also marked in red) in the goodness-of-fit test (Stage II), while \flower~denotes the effective scaling law from Stage III.}
\label{tab:beam_unieval}
\end{table*}

\begin{figure*}[h]
    \centering
    \includegraphics[width=\textwidth]{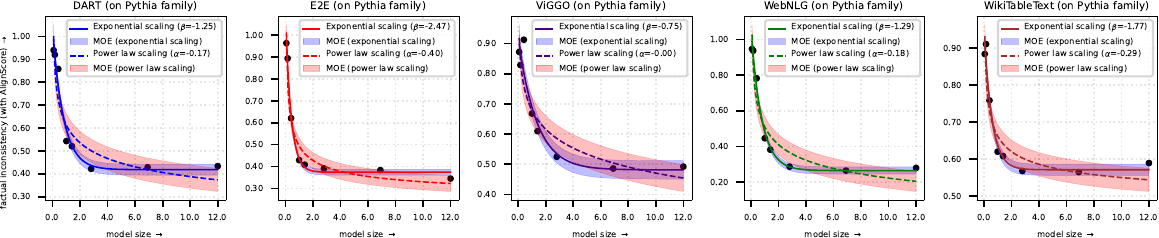}\\
    \vspace{2mm}
    \includegraphics[width=\textwidth]{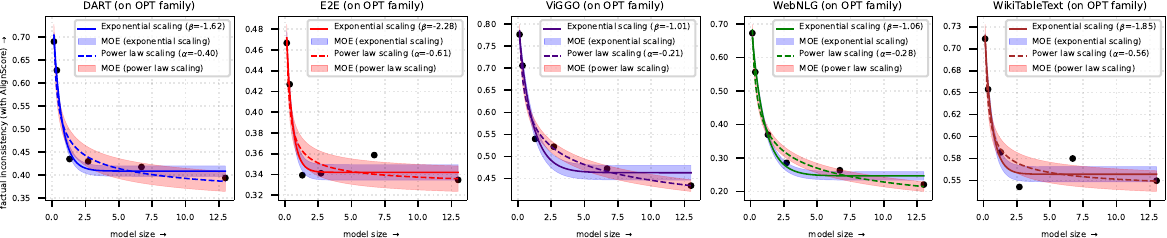}\\
    \vspace{2mm}
    \includegraphics[width=\textwidth]{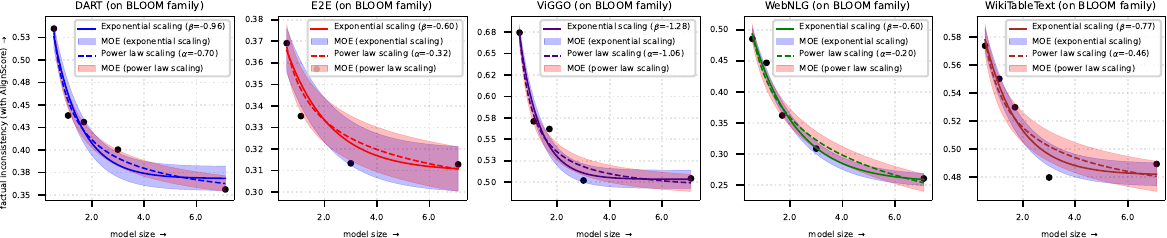}
    \caption{Visualization of exponential and power law scaling of factual inconsistency (\textsc{AlignScore}) across datasets and LLM families, with the margin of error (MOE) and $95\%$ confidence intervals on residuals. Texts generated using the \ul{topk search decoding algorithm}.}
    \label{fig:fit_alignscore_topk}
\end{figure*}

\begin{table*}[h]
\resizebox{\textwidth}{!}{
\begin{tabular}{@{}cc|ccccc|ccccc@{}}
\toprule
\multicolumn{2}{c|}{} & \multicolumn{5}{c|}{Results of stage I} & \multicolumn{5}{c}{Results of stage II and III} \\ \midrule
LLM family & Scaling law & DART & E2E & ViGGO & WebNLG & WikiTableText & DART & E2E & ViGGO & WebNLG & WikiTableText \\ \midrule
 & Exponential & ${1.70}\text{e-}{03}$ & ${1.74}\text{e-}{03}$ & ${4.01}\text{e-}{04}$ & ${2.90}\text{e-}{04}$ & ${2.39}\text{e-}{03}$ & \cellcolor[HTML]{E67C73}\xmark & \cellcolor[HTML]{E67C73}\xmark & \cmark & \cmark (\flower) & \cmark \\
\multirow{-2}{*}{BLOOM} & Power law & ${4.85}\text{e-}{04}$ & ${1.25}\text{e-}{04}$ & ${3.89}\text{e-}{04}$ & ${4.50}\text{e-}{03}$ & ${3.91}\text{e-}{04}$ & \cmark & \cellcolor[HTML]{E67C73}\xmark & \cellcolor[HTML]{FFD666}\cmark (\flower) & \cmark & \cellcolor[HTML]{E67C73}\xmark \\ \midrule
 & Exponential & ${8.14}\text{e-}{04}$ & ${3.97}\text{e-}{03}$ & ${1.15}\text{e-}{03}$ & ${6.12}\text{e-}{04}$ & ${3.80}\text{e+}{02}$ & \cmark (\flower) & \cmark (\flower) & \cmark (\flower) & \cmark (\flower) & \cmark (\flower) \\
\multirow{-2}{*}{OPT} & Power law & ${4.47}\text{e-}{03}$ & ${2.84}\text{e-}{03}$ & ${9.17}\text{e-}{04}$ & ${3.69}\text{e-}{03}$ & ${4.28}\text{e-}{03}$ & \cmark & \cmark & \cmark & \cmark & \cmark \\ \midrule
 & Exponential & ${5.76}\text{e-}{04}$ & ${6.57}\text{e-}{04}$ & ${2.31}\text{e-}{03}$ & ${1.28}\text{e-}{03}$ & ${8.77}\text{e-}{04}$ & \cmark (\flower) & \cmark (\flower) & \cmark (\flower) & \cmark (\flower) & \cmark (\flower) \\
\multirow{-2}{*}{Pythia} & Power law & ${9.31}\text{e-}{03}$ & ${1.31}\text{e-}{02}$ & ${7.78}\text{e-}{03}$ & ${1.36}\text{e-}{02}$ & ${1.52}\text{e-}{03}$ & \cmark & \cmark & \cmark & \cmark & \cmark \\ \bottomrule
\end{tabular}}
\caption{\ul{[Case: top-k decoding]} Results of the validation framework (all three stages) for exponential and power law scaling of factual inconsistency (\textsc{AlignScore}). High held-out losses (Stage I) are highlighted in red. \cmark/\xmark~indicates pass/fail (also marked in red) in the goodness-of-fit test (Stage II), while \flower~denotes the effective scaling law from Stage III.}
\label{tab:topk_alignscore}
\end{table*}

\begin{figure*}[h]
    \centering
    \includegraphics[width=\textwidth]{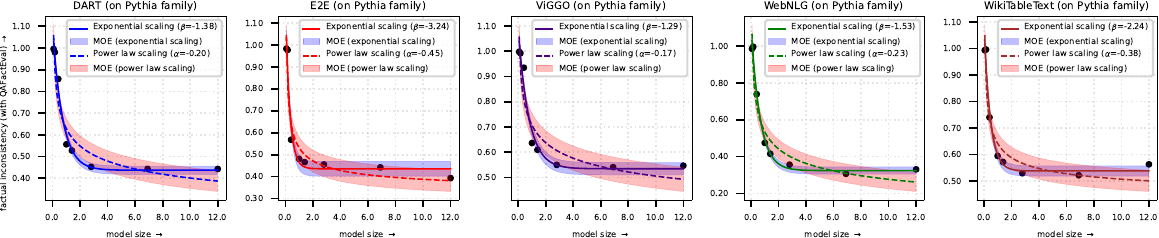}\\
    \vspace{2mm}
    \includegraphics[width=\textwidth]{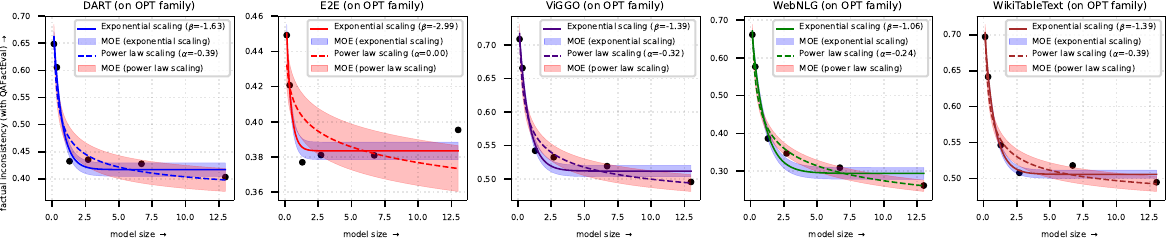}\\
    \vspace{2mm}
    \includegraphics[width=\textwidth]{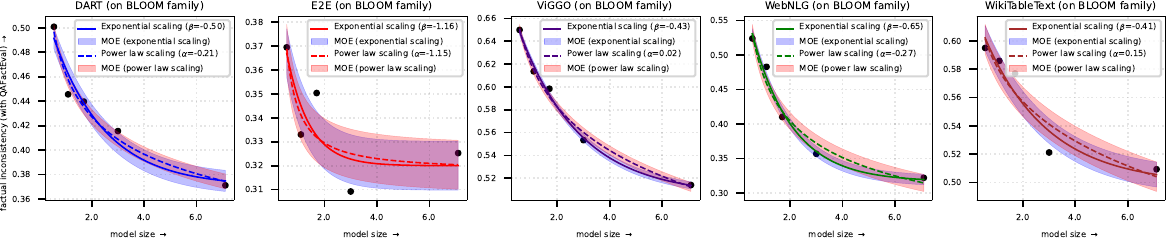}
    \caption{Visualization of exponential and power law scaling of factual inconsistency (\textsc{QAFactEval}) across datasets and LLM families, with margin of error (MOE) and $95\%$ confidence intervals on residuals. Texts generated using the \ul{topk search decoding algorithm}.}
    \label{fig:fit_qafacteval_topk}
\end{figure*}

\begin{table*}[h]
\resizebox{\textwidth}{!}{
\begin{tabular}{@{}cc|ccccc|ccccc@{}}
\toprule
\multicolumn{2}{c|}{} & \multicolumn{5}{c|}{Results of stage I} & \multicolumn{5}{c}{Results of stage II and III} \\ \midrule
LLM family & Scaling law & DART & E2E & ViGGO & WebNLG & WikiTableText & DART & E2E & ViGGO & WebNLG & WikiTableText \\ \midrule
 & Exponential & ${8.47}\text{e-}{04}$ & ${2.77}\text{e-}{04}$ & ${2.86}\text{e-}{05}$ & ${1.80}\text{e-}{04}$ & ${1.34}\text{e-}{03}$ & \cellcolor[HTML]{E67C73}\xmark & \cellcolor[HTML]{E67C73}\xmark & \cmark (\flower) & \cmark (\flower) & \cellcolor[HTML]{E67C73}\xmark \\
\multirow{-2}{*}{BLOOM} & Power law & ${3.52}\text{e-}{04}$ & ${3.60}\text{e-}{04}$ & ${3.73}\text{e-}{05}$ & ${5.46}\text{e-}{04}$ & ${7.83}\text{e-}{04}$ & \cmark & \cellcolor[HTML]{E67C73}\xmark & \cmark & \cmark & \cellcolor[HTML]{E67C73}\xmark \\ \midrule
 & Exponential & ${1.54}\text{e-}{03}$ & ${5.44}\text{e-}{04}$ & ${3.86}\text{e-}{04}$ & ${1.17}\text{e-}{03}$ & ${6.48}\text{e-}{05}$ & \cmark (\flower) & \cmark & \cmark (\flower) & \cmark & \cmark (\flower) \\
\multirow{-2}{*}{OPT} & Power law & ${1.37}\text{e-}{03}$ & ${8.75}\text{e-}{05}$ & ${1.72}\text{e-}{03}$ & ${6.90}\text{e-}{03}$ & ${2.49}\text{e+}{01}$ & \cmark & \cellcolor[HTML]{E67C73}\xmark & \cmark & \cellcolor[HTML]{FFD666}\cmark (\flower) & \cmark \\ \midrule
 & Exponential & ${1.50}\text{e-}{03}$ & ${1.48}\text{e-}{02}$ & ${1.55}\text{e-}{03}$ & ${2.76}\text{e-}{03}$ & ${1.48}\text{e-}{03}$ & \cmark (\flower) & \cmark (\flower) & \cmark (\flower) & \cmark (\flower) & \cmark (\flower) \\
\multirow{-2}{*}{Pythia} & Power law & ${2.30}\text{e-}{02}$ & ${1.84}\text{e-}{01}$ & ${5.73}\text{e-}{03}$ & ${4.43}\text{e-}{02}$ & ${3.10}\text{e-}{03}$ & \cmark & \cmark & \cmark & \cmark & \cmark \\ \bottomrule
\end{tabular}}
\caption{\ul{[Case: top-k decoding]} Results of the validation framework (all three stages) for exponential and power law scaling of factual inconsistency (\textsc{QAFactEval}). High held-out losses (Stage I) are highlighted in red. \cmark/\xmark~indicates pass/fail (also marked in red) in the goodness-of-fit test (Stage II), while \flower~denotes the effective scaling law from Stage III.}
\label{tab:topk_qafacteval}
\end{table*}

\begin{figure*}[h]
    \centering
    \includegraphics[width=\textwidth]{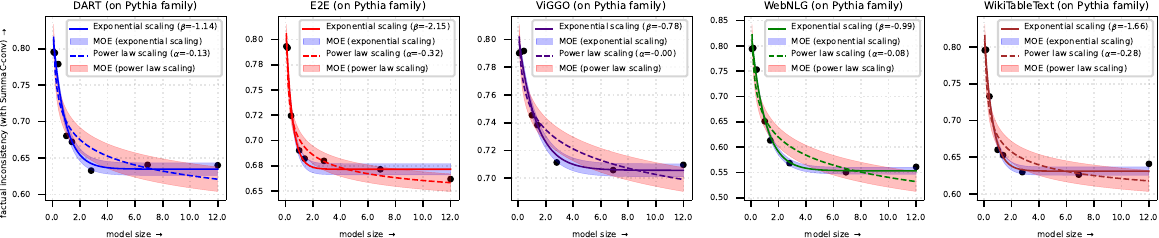}\\
    \vspace{2mm}
    \includegraphics[width=\textwidth]{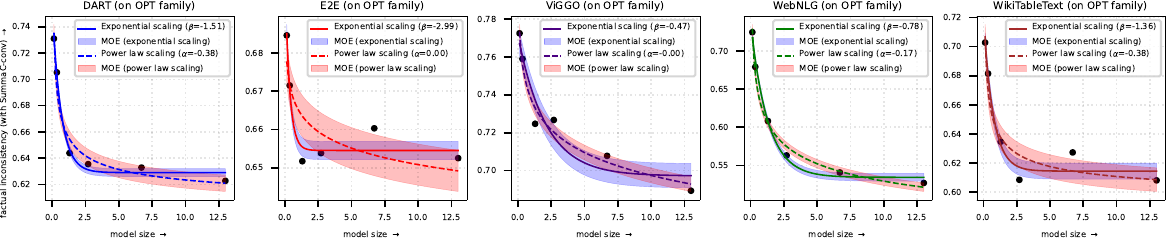}\\
    \vspace{2mm}
    \includegraphics[width=\textwidth]{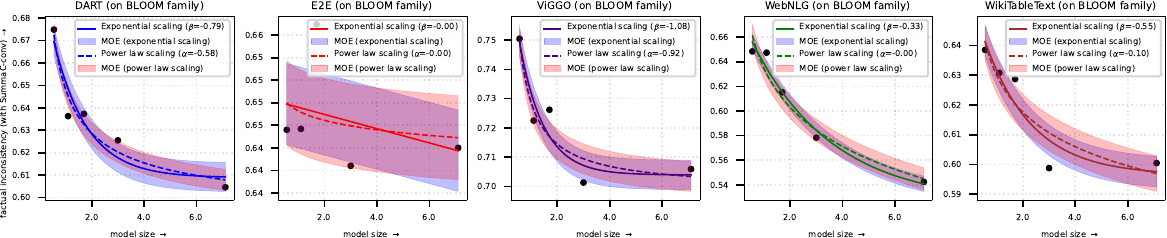}
    \caption{Visualization of exponential and power law scaling of factual inconsistency (\textsc{SummaC-conv}) across datasets and LLM families, with margin of error (MOE) and $95\%$ confidence intervals on residuals. Texts generated using the \ul{topk search decoding algorithm}.}
    \label{fig:fit_summac_topk}
\end{figure*}

\begin{table*}[h]
\resizebox{\textwidth}{!}{
\begin{tabular}{@{}cc|ccccc|ccccc@{}}
\toprule
\multicolumn{2}{c|}{} & \multicolumn{5}{c|}{Results of stage I} & \multicolumn{5}{c}{Results of stage II and III} \\ \midrule
LLM family & Scaling law & DART & E2E & ViGGO & WebNLG & WikiTableText & DART & E2E & ViGGO & WebNLG & WikiTableText \\ \midrule
 & Exponential & ${2.13}\text{e-}{04}$ & ${4.52}\text{e-}{05}$ & ${7.56}\text{e-}{05}$ & ${1.68}\text{e-}{03}$ & ${8.72}\text{e-}{04}$ & \cellcolor[HTML]{E67C73}\xmark & \cellcolor[HTML]{E67C73}\xmark & \cellcolor[HTML]{E67C73}\xmark & \cmark & \cellcolor[HTML]{E67C73}\xmark \\
\multirow{-2}{*}{BLOOM} & Power law & ${4.90}\text{e-}{05}$ & ${2.76}\text{e-}{05}$ & ${8.40}\text{e-}{05}$ & ${1.48}\text{e-}{04}$ & ${1.60}\text{e-}{04}$ & \cmark & \cellcolor[HTML]{E67C73}\xmark & \cellcolor[HTML]{E67C73}\xmark & \cellcolor[HTML]{E67C73}\xmark & \cellcolor[HTML]{E67C73}\xmark \\ \midrule
 & Exponential & ${3.00}\text{e-}{05}$ & ${4.87}\text{e-}{05}$ & ${1.82}\text{e-}{04}$ & ${1.34}\text{e-}{04}$ & ${4.44}\text{e+}{02}$ & \cmark (\flower) & \cmark & \cmark (\flower) & \cmark (\flower) & \cmark (\flower) \\
\multirow{-2}{*}{OPT} & Power law & ${1.11}\text{e-}{03}$ & ${7.14}\text{e-}{05}$ & ${6.31}\text{e-}{05}$ & ${1.72}\text{e-}{04}$ & ${8.68}\text{e-}{04}$ & \cmark & \cellcolor[HTML]{E67C73}\xmark & \cmark & \cmark & \cmark \\ \midrule
 & Exponential & ${9.16}\text{e-}{04}$ & ${7.38}\text{e-}{05}$ & ${7.46}\text{e-}{05}$ & ${5.39}\text{e-}{05}$ & ${1.48}\text{e-}{04}$ & \cmark (\flower) & \cmark (\flower) & \cmark (\flower) & \cmark (\flower) & \cmark (\flower) \\
\multirow{-2}{*}{Pythia} & Power law & ${1.08}\text{e-}{03}$ & ${2.50}\text{e-}{04}$ & ${7.98}\text{e-}{05}$ & ${2.99}\text{e-}{02}$ & ${3.12}\text{e-}{04}$ & \cmark & \cmark & \cmark & \cmark & \cmark \\ \bottomrule
\end{tabular}}
\caption{\ul{[Case: top-k decoding]} Results of the validation framework (all three stages) for exponential and power law scaling of factual inconsistency (\textsc{SummaC-conv}). High held-out losses (Stage I) are highlighted in red. \cmark/\xmark~indicates pass/fail (also marked in red) in the goodness-of-fit test (Stage II), while \flower~denotes the effective scaling law from Stage III.}
\label{tab:topk_summac}
\end{table*}

\begin{figure*}[h]
    \centering
    \includegraphics[width=\textwidth]{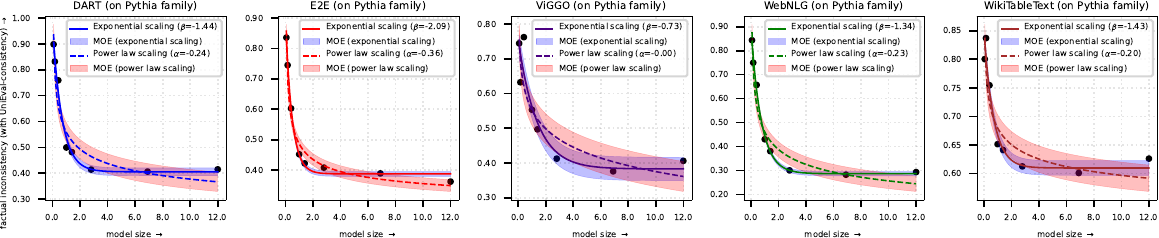}\\
    \vspace{2mm}
    \includegraphics[width=\textwidth]{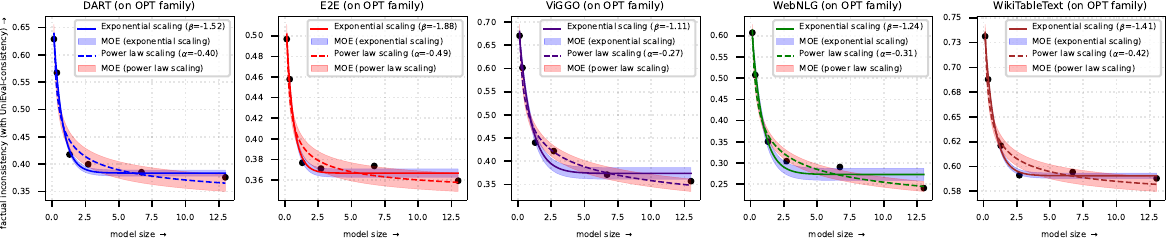}\\
    \vspace{2mm}
    \includegraphics[width=\textwidth]{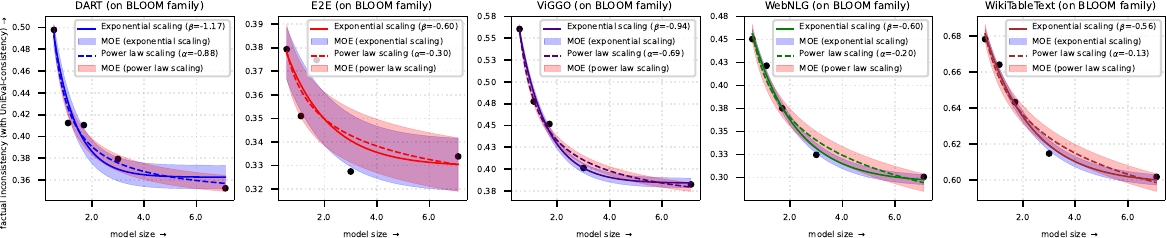}
    \caption{Visualization of exponential and power law scaling of factual inconsistency (\textsc{UniEval-fact}) across datasets and LLM families, with margin of error (MOE) and $95\%$ confidence intervals on residuals. Texts generated using the \ul{topk search decoding algorithm}.}
    \label{fig:fit_unieval_topk}
\end{figure*}

\begin{table*}[h]
\resizebox{\textwidth}{!}{
\begin{tabular}{@{}cc|ccccc|ccccc@{}}
\toprule
\multicolumn{2}{c|}{} & \multicolumn{5}{c|}{Results of stage I} & \multicolumn{5}{c}{Results of stage II and III} \\ \midrule
LLM family & Scaling law & DART & E2E & ViGGO & WebNLG & WikiTableText & DART & E2E & ViGGO & WebNLG & WikiTableText \\ \midrule
 & Exponential & ${4.84}\text{e-}{04}$ & ${1.94}\text{e-}{03}$ & ${1.99}\text{e-}{04}$ & ${5.07}\text{e-}{05}$ & ${1.03}\text{e-}{05}$ & \cellcolor[HTML]{E67C73}\xmark & \cellcolor[HTML]{E67C73}\xmark & \cmark & \cmark (\flower) & \cmark (\flower) \\
\multirow{-2}{*}{BLOOM} & Power law & ${7.37}\text{e-}{04}$ & ${9.46}\text{e-}{05}$ & ${1.55}\text{e-}{04}$ & ${2.52}\text{e-}{04}$ & ${1.09}\text{e-}{03}$ & \cmark & \cellcolor[HTML]{E67C73}\xmark & \cellcolor[HTML]{FFD666}\cmark (\flower) & \cmark & \cmark \\ \midrule
 & Exponential & ${2.05}\text{e-}{04}$ & ${5.95}\text{e-}{05}$ & ${5.23}\text{e-}{04}$ & ${7.91}\text{e-}{04}$ & ${8.86}\text{e-}{04}$ & \cmark (\flower) & \cmark (\flower) & \cmark & \cmark (\flower) & \cmark (\flower) \\
\multirow{-2}{*}{OPT} & Power law & ${7.80}\text{e-}{03}$ & ${3.85}\text{e-}{04}$ & ${2.05}\text{e-}{03}$ & ${2.98}\text{e-}{03}$ & ${5.55}\text{e-}{04}$ & \cmark & \cmark & \cellcolor[HTML]{FFD666}\cmark (\flower) & \cmark & \cmark \\ \midrule
 & Exponential & ${5.08}\text{e-}{04}$ & ${7.40}\text{e-}{05}$ & ${6.71}\text{e-}{03}$ & ${1.47}\text{e-}{04}$ & ${5.26}\text{e-}{04}$ & \cmark (\flower) & \cmark (\flower) & \cmark (\flower) & \cmark (\flower) & \cmark (\flower) \\
\multirow{-2}{*}{Pythia} & Power law & ${7.34}\text{e-}{03}$ & ${6.73}\text{e-}{04}$ & ${1.54}\text{e-}{03}$ & ${3.71}\text{e-}{03}$ & ${5.66}\text{e-}{03}$ & \cmark & \cmark & \cmark & \cmark & \cmark \\ \bottomrule
\end{tabular}}
\caption{\ul{[Case: top-k decoding]} Results of the validation framework (all three stages) for exponential and power law scaling of factual inconsistency (\textsc{UniEval-fact}). High held-out losses (Stage I) are highlighted in red. \cmark/\xmark~indicates pass/fail (also marked in red) in the goodness-of-fit test (Stage II), while \flower~denotes the effective scaling law from Stage III.}
\label{tab:topk_unieval}
\end{table*}

\end{document}